\documentclass[10pt,twocolumn,twoside]{IEEEtran}

\usepackage{mathtools}
\usepackage{tcolorbox}
\usepackage{color}
\usepackage{theorem}
\usepackage{times,amsmath,epsfig}
\usepackage{amssymb}
\usepackage{caption}
\usepackage{subcaption}
\usepackage{cite}
\usepackage{cases}
\usepackage{url}
\usepackage{algorithmic}
\usepackage{algorithm}
\usepackage{needspace}

\usepackage{tikz}
\usetikzlibrary{shapes,arrows}
\input{mysymbol.sty}
\tikzstyle{phantom vertex} = [ ellipse, 
                               anchor = center, 
                               minimum height = 1*\unit, 
                               minimum width  = 1*\unit,
                               inner sep=0pt,
                               anchor=center]
\tikzstyle{red vertex}   = [black, fill = red!20,   phantom vertex, draw]
\tikzstyle{black vertex} = [black, fill = black!20, phantom vertex, draw]
\tikzstyle{blue vertex}  = [black, fill = blue!20,  phantom vertex, draw]
\tikzstyle{green vertex} = [black, fill = green!20,  phantom vertex, draw]
\tikzstyle{yellow vertex} = [black, fill = yellow!20,  phantom vertex, draw]
\tikzstyle{cyan vertex} = [black, fill = cyan!20,  phantom vertex, draw]
\tikzstyle{vertex}       = [draw, phantom vertex]

\tikzstyle{point} = [ellipse, inner sep=0pt, draw, fill=white, anchor = center,
                     minimum height = 0.05*\unit, minimum width  = 0.05*\unit]

\newcommand{\vect}{\mathrm{vec}}

\newtheorem{mytheorem}{\bf Theorem}

\newtheorem{mylemma}{\bf Lemma}

\newtheorem{problem}{\bf Problem}
\newtheorem{myclaim}{\bf Claim}

\renewcommand{\baselinestretch}{0.98}
\newtcolorbox{myblockt}[1]{colback=urblue!5!white,
	colframe=urblue,fonttitle=\bfseries,
	title=#1}

\newtcolorbox{myblock}{colback=urblue!5!white,
	colframe=urblue,fonttitle=\bfseries}

\title{Joint Inference of Multiple Graphs from Matrix Polynomials}

\author{\IEEEauthorblockN{Madeline Navarro, Yuhao Wang, Antonio G. Marques, Caroline Uhler, and Santiago Segarra}
	\thanks{ M. Navarro and S. Segarra are with the Dept. of ECE, Rice University. Y. Wang is with IIIS, Tsinghua University and Shanghai Qi Zhi Institute. A. G. Marques is with the Dept. of ST \& Comm., King Juan Carlos University. C. Uhler is with the Dept. of EECS and IDSS, MIT.
	Research was supported by NSF (DMS-1651995 and CCF-2008555), and the Spanish Federal grants KLINILYCS (TEC2016-75361-
R) and SPGraph (PID2019-105032GB-I00).
	Emails: madeline.navarro@rice.edu, yuhaow@tsinghua.edu.cn, antonio.garcia.marques@urjc.es, cuhler@mit.edu, segarra@rice.edu.
	Preliminary results appeared in a conference publication~\cite{segarra_2017_joint}.
		}
}
\begin{document}
\maketitle

%%%%%%%%%%%%%%%%
\begin{abstract}%
Inferring graph structure from observations on the nodes is an important and popular network science task. 
Departing from the more common inference of a single graph and motivated by social and biological networks, we study the problem of jointly inferring multiple graphs from the observation of signals at their nodes (graph signals), which are assumed to be stationary in the sought graphs.
From a mathematical point of view, graph stationarity implies that the mapping between the covariance of the signals and the sparse matrix representing the underlying graph is given by a matrix polynomial. A prominent example is that of Markov random fields, where the inverse of the covariance yields the sparse matrix of interest. From a modeling perspective, stationary graph signals can be used to model linear network processes evolving on a set of (not necessarily known) networks. Leveraging that matrix polynomials commute, a convex optimization method along with sufficient conditions that guarantee the recovery of the true graphs are provided when perfect covariance information is available.
Particularly important from an empirical viewpoint, we provide high-probability bounds on the recovery error as a function of the number of signals observed and other key problem parameters.
Numerical experiments using synthetic and real-world data demonstrate the effectiveness of the proposed method with perfect covariance information as well as its robustness in the noisy regime.
 \end{abstract}
%%%%%%%%%%%%%%%%

%%%%%%%%%%%%%%%%
%%%%%%%%%%%%%%%%

%%%%%%%%%%%%%%%%%%%%%%%%%%%%%%%%%%%%%%%%%%%%%%%%%%%%%%%%%%%%%%
% INTRODUCTION
\section{Introduction}\label{S:Introduction}

Inferring the topology of a network (graph) from a set of nodal observations is a prominent problem in statistics, network science, machine learning, and signal processing (SP)~\cite{kolaczyk2009book,sporns2012book}, with applications including power, communications, and brain networks \cite{ortega_2018_graph, marques2020editorial, djuric2018cooperative}, to name a few. 
Networks can exist as actual physical entities or can be convenient mathematical \textit{representations} describing \textit{parsimonious pairwise relationships} between data. 
Transversal to the particularities of the setup, the fundamental assumption in these network-inference approaches is the formalization of a relation between the topology of the sought network and the properties of the nodal observations. Notable approaches include correlation networks~\cite[Ch. 7.3.1]{kolaczyk2009book}, partial correlations and (Gaussian) Markov random fields~\cite{meinshausen06,GLasso2008,kolaczyk2009book,Lake10discoveringstructure}, structured equation models~\cite{BazerqueGeneNetworks,BainganaInfoNetworks}, graph-SP-based approaches~\cite{mateos_2019_connecting, dong_2019_learning, MeiGraphStructure, DongLaplacianLearning, Kalofolias2016inference_smoothAISTATS16, pavez_laplacian_inference_icassp16, segarra2017network, pasdeloup2017characterization}, as well as their non-linear generalizations~\cite{Karanikolas_icassp16,shen2017kernel}. 

While most of the existing works have looked at the problem of identifying a single network, many contemporary setups involve \textit{multiple} related networks, each of them with a subset of available observations. 
Examples of this multi-graph setup arise in multi-hop communication networks deployed in \textit{dynamic} environments where links are created or destroyed as nodes change their position, in brain analytics where observations for different patients are available and the objective is to estimate their brain functional networks, in gene-to-gene networks where the goal is to identify pairwise interactions between genes and measurements for different tissues, or in social networks where the same set of users can have different types of social interactions~\cite{arroyo2019inference,murase2014multilayer,bindu2017discovering}. 
Arguably, in many contemporary applications, dealing with multiple networks may be more the rule than the exception. 
Last but not least, one must also note that the joint identification of multiple graphs can be useful even if the interest is only in one of the networks, since joint formulations exploit additional sources of information and, hence, are likely to give rise to better solutions. 

Given the previous motivations, our goal in this paper is to develop new \textit{schemes for the joint inference of multiple networks} that build on recent results from graph SP (GSP) and, in particular, on the notion of graph stationarity~\cite{marques2017stationary, perraudin2017stationary, girault2015translation}. 
In the last years, GSP has emerged as a way to generalize tools originally conceived to process signals with regular supports (time or space) to signals defined in heterogeneous domains represented by graphs~\cite{ortega_2018_graph}. 
The systematic approach put forth relies on the definition of a \textit{graph shift operator} (GSO), which is a \textit{sparse square matrix} capturing the local interactions (connections) between pairs of nodes. Within the GSP framework, the GSO constitutes the basic signal operator in the vertex domain, and its eigenvectors define the \textit{graph} Fourier transform, which enables the analysis and processing of graph signals in a proper frequency domain. 
The GSO (typically assumed to have the form of an adjacency or Laplacian matrix) is also critical to define the notion of graph stationarity~\cite{marques2017stationary, perraudin2017stationary, girault2015translation}, which generalizes the classical notion of time-stationarity to signals defined on graphs and constitutes the fundamental GSP concept utilized in this paper. 
Given the covariance matrix associated with a random graph process, \textit{graph stationarity} requires this covariance and the GSO representing the support of the process to have the same eigenvectors.
This requirement, which is equivalent to saying that there exists a \textit{polynomial mapping} between the \textit{sparse shift} and the \textit{covariance} matrix, is fairly general, encompassing classical approaches such as correlation and conditional independent networks \cite{mateos_2019_connecting}.

Leveraging those concepts, we can now describe more concretely the GSP-inspired approach put forth in this paper, which aims at inferring the topology of the multiple networks by solving an optimization problem where we look for graph shift matrices that are sparse, guarantee that the observed signals are stationary on the identified graphs, and force the different shifts to be close to each other according to a pre-specified level of similarity. 
Our formulation also takes into account additional structural information that may be available (such as the GSOs corresponding to a particular type of Laplacian, or being an adjacency matrix without self-loops). 
Together with the novel approach for the formulation of the joint topology inference of multiple networks, the paper also identifies theoretical conditions under which convex relaxations are able to find the optimal sparse structure in noiseless settings (Theorem~\ref{T:theo_perfect_recovery}) as well as a detailed theoretical analysis of the probability of robust recovery in the more practical noisy scenario (Theorem~\ref{T:robust_recovery}).

%%%%%%%%%%%%%%%%
\medskip\noindent\textbf{Related work.} 
Although noticeably less than its single-network counterpart, joint inference of multiple networks (a structure oftentimes referred to as a multi-layer graph~\cite{oselio_hero2014multilayer,sardellitti2019enabling}) has attracted attention for different versions of the problem. The most widely studied one is that of inferring (tracking) the topology of time-varying networks. The standard approach is to assume that the variation is smooth across time, so that the graph-inference problem is regularized with a term that promotes changes between consecutive graphs to be small in some pre-specified norm ~\cite{zhou2010time,kalofolias2017learning,baingana2017tracking,yamada2019time,kao2017disc}. 
A second cluster of works focuses on the joint inference of multiple (Gaussian) Markov random fields without assuming a temporal dimension~\cite{guo2011joint,danaher2014joint,ryali2012estimation,honorio2010multi,varoquaux2010brain,gan2019bayesian,cai2016joint,wang_2020_high}. Each graph has its own subset of signal observations, and the goal is the joint recovery of sparse precision matrices. The formulated problems typically correspond to generalizations of the graphical lasso formulation, accounting for either similarity or common structure across the multiple graphs.
A third class of more involved approaches looks at the case of signal mixtures, where the assignment of each graph to observed signals is unknown; see, e.g. \cite{lotsi2013high,hao2017simultaneous} for sparse precision-matrices approaches and  \cite{araghi2019k,maretic2020graphLaplacianmixture} for GSP-based ones. In those cases, not only the graphs but also the signal-to-graph assignments must be inferred. 
This results in recovery problems that are more challenging to solve, with Gaussianity being often assumed to leverage expectation-maximization approaches. 
In most cases, the focus is on the problem formulation and algorithmic design, without characterizing the recovery performance theoretically. 
%Additional past analyses of multiple graphs outside those three categories also exist. For example, \cite{arroyo2019inference} looks at the development of multiple network models to represent heterogeneous graphs that share certain characteristics (such as vertex set or invariant subspaces) and \cite{levin2017central} presents a method to estimate latent positions of multiple graphs. \blue{[AGM: I am not a big fan of the two previous sentences.]} 
The present paper is more closely related to the second cluster of works but goes beyond sparse precision matrices and provides novel theoretical guarantees, as detailed next.

%%%%%%%%%%%%%%%%
\medskip\noindent\textbf{Contributions.} 
{This paper's contributions are fourfold:\\
i) We propose an efficient optimization-based solution to the problem of joint inference of sparse graphs from the observation of stationary graph signals.\\
ii) We determine sufficient conditions under which the proposed efficient method is guaranteed to recover the underlying set of true sparse graphs (Theorem~\ref{T:theo_perfect_recovery}).\\
iii) We show the robustness of our method by deriving tight high-probability upper bounds on the recovery errors when imperfect covariance information is used to solve the joint inference problem (Theorem~\ref{T:robust_recovery}).\\
iv) We rely on both synthetic and real-world data to compare the performance of joint and separate inference, validate the conditions for guaranteed recovery, and demonstrate the robustness of the proposed method in noisy settings.
%Relative to our work in \cite{segarra2017network}, here we look at multiple shifts and bypass the need for finding the eigenvalues of the shifts, giving rise to simpler optimization problems. Relative to the generalization of graphical lasso approaches to handle multiple shifts \cite{guo2011joint,danaher2014joint, ryali2012estimation,honorio2010multi,varoquaux2010brain}, we consider generic graph-distance functions and allow for different levels of proximity among graphs. More importantly, we assume a more general polynomial mapping between the GSO and the covariance matrix. This mapping includes as particular cases correlation networks, sparse precision networks, some types of structural equation models, as well as more involved higher order relations (e.g., polynomial Laplacian kernels) not considered before.
%%%%%%%%%%%%%%%%

%%%%%%%%%%%%%%%%
\medskip\noindent\textbf{Paper outline.} 
The remainder of this paper is organized as follows. 
In Section \ref{S:prelim_problem}, we describe the main problem and the required assumptions, introduce some background on signal stationarity, and discuss graph similarity notions. 
Section \ref{S:ConvexRelax_RecovGuarantess} first introduces the non-convex problem of jointly inferring multiple graphs given covariance matrices. 
While true covariances are not often available, this problem lays the foundation for more realistic problem setups. 
The inference problem is further developed in Section \ref{Ss:relaxation}, where we introduce the convex relaxation of the sparse graph learning problem and show conditions that lead to perfect recovery. 
In Section \ref{S:robust_recovery}, we demonstrate the robustness of our method when only noisy or imperfect covariance matrices are available, and we provide a novel bound on the recovery error.
Through experiments on synthetic and real-world data, we illustrate the performance of the proposed joint graph inference method in Section~\ref{S:numerical_experiments}. Finally, we discuss conclusions and possible future research directions in Section~\ref{S:conclusions}.\footnote{\textbf{Notation:}
The entries of a matrix $\mathbf{X}$ and a (column) vector $\mathbf{x}$ are denoted by $X_{ij}$ and $x_i$, respectively.  The notation $^\top$ and $^\dag$ stands for transpose and pseudo-inverse, respectively. With the size clear from the context, $\mathbf{0}$ and $\mathbf{1}$ refer to the all-zero and all-one vectors, and $\bbe_i$ refers to the $i$-th canonical vector, i.e., a vector whose entries are all zero except the $i$-th one, which is set to one. Sets are represented by calligraphic capital letters. Given an implicit set $\ccalB$ and a set $\ccalA \subseteq \ccalB$, the set $\ccalA^c$ stands for the complement set of $\ccalA$, i.e., $\ccalA^c=\ccalB \setminus \ccalA $ contains the elements in $\ccalB$ that do not belong to $\ccalA$. Moreover, $\bbX_{\ccalI}$ denotes a submatrix of $\bbX$ formed by selecting the rows of $\bbX$ indexed by $\ccalI$. The expression $\bbX^\top_\ccalI$ denotes first selecting the rows and then transposing, whereas $[\bbX^\top]_\ccalI$ is used to denote the opposite order of operations. For a vector $\bbx$, $\diag(\mathbf{x})$ is a diagonal matrix whose $i$-th diagonal entry is $x_i$; when applied to a matrix, $\diag(\bbX)$ is a vector with the diagonal elements of $\bbX$. The vertical concatenation of the columns of $\bbX$ is denoted as $\vect(\bbX)$.
The operators $\circ$, $\otimes$, and $\odot$ stand for the Hadamard (element-wise), Kronecker, and Khatri-Rao (column-wise Kronecker) matrix products, while the operator $\oplus$ denotes the Kronecker matrix sum, so that $\bbX\oplus\bbY=\bbX\otimes\bbI+\bbI\otimes\bbY$, where the size of each of the identity matrices is chosen to make the dimensions of the matrices consistent. 
$\| \bbX \|_p$  is the matrix norm induced by the vector $\ell_p$ norm, not to be confused with  $\| \textrm{vec}(\bbX) \|_p $. 
$\mathrm{ker}(\bbX)$ and $\mathrm{Im}(\bbX)$ refer to the null space and the span of the columns of $\bbX$, respectively.
The notation $O(\cdot)$ and $o(\cdot)$ entail the usual asymptotic meaning and we write that $f \asymp g$ if $f = O(g)$ and $g = O(f)$.}
%
%After surveying the required GSP background, Section \ref{S:prelim_problem} states the problem of joint topology inference at a general level and then formulates it in a rigorous optimization form. The implications of the assumptions are discussed in detail and the different parts of the optimization problem are analyzed. Since the formulation may include sparsity-related terms which are non-convex, Section \ref{S:ConvexRelax_RecovGuarantess} presents convex relaxations with probable guarantees,  establishing also that the proposed algorithm can identify the underlying network topologies robustly. The focus of Section \ref{S:Reduced-complexity_algorithms} is on developing algorithms with reduced complexity that allow for carrying out the inference in a computationally efficient manner. Emphasis in those two sections is laid on GSOs representing adjacency matrices, but the methodology and corresponding findings can be generalized to other matrix representations of graphs, as discussed in more detail in the paper. In Section \ref{S:Simulations}, numerical tests corroborate our theoretical findings and confirm that the novel approach compares favorably with respect to graphical lasso methods and recent graph signal processing-based topology inference algorithms. Test cases include the recovery of gene and structural brain networks, based on real data \cite{}.   %Concluding remarks are given in Section \ref{S:Conclusions}.
%%%%%%%%%%%%%%%%
%
%%%%%%%%%%%%%%%%
%\\ \medskip\noindent\textbf{Notation.}{

%%%%%%%%%%%%%%%%

%%%%%%%%%%%%%%%%
\subsection{Fundamentals of graph signal processing} \label{Ss:fund_GSP}

Let us consider a generic weighted and undirected graph $\ccalG$ consisting of a node set $\ccalN$ of known cardinality $N$, an edge set $\ccalE$ of unordered pairs of elements in $\ccalN$, and edge weights $A_{ij}\in\reals$ such that $A_{ij}=A_{ji}\neq 0$ for all $(i,j)\in\ccalE$. The edge weights $A_{ij}$ are collected as entries of the symmetric adjacency matrix $\bbA$ and the node degrees in the diagonal matrix $\bbD:=\diag(\bbA\bbone)$. These are used to form the combinatorial Laplacian matrix $\bbL_c:=\bbD-\bbA$ and the normalized Laplacian $\bbL:=\bbI - \bbD^{-1/2} \bbA\bbD^{-1/2}$. More broadly, one can define a generic GSO $\bbS\in\reals^{N\times N}$ as any matrix whose off-diagonal sparsity pattern is equal to that of the adjacency matrix of $\ccalG$~\cite{SandryMouraSPG_TSP13}. Although the choice of $\bbS$ can be adapted to the problem at hand, most existing works set it to either $\bbA$, $\bbL_c$, or $\bbL$. If the GSO is symmetric, its normal eigendecomposition $\bbS=\bbV\bbLambda\bbV^\top$, with $\bbV$ unitary and $\bbLambda$ diagonal, exists. Suppose now that we associate a value (observation) with each node of the graph. Those $N$ values form a graph signal that can be conveniently represented as the vector $\bbx=[x_1,...,x_N]^\top \in\mbR^N$, with entry $x_n$ denoting the signal value at node $n$. 
A key aspect when dealing with graph signals is the definition of  meaningful operators able to relate different signals while efficiently accounting for the topology of the graph. 
Linear graph filters, which are defined as $\bbH=\sum_{l=0}^{\infty} h_l \bbS^l$, i.e., matrix polynomials of the GSO \cite{SandryMouraSPG_TSP13}, are the most widely-adopted alternative.
Graph filters have shown to be useful not only to process graph signals (e.g., used for denoising and interpolation), but also to model linear network dynamics and network processes \cite{djuric2018cooperative}. To illustrate this latter point,  consider a dynamic network setup where the initial state (value) of most nodes is zero and only a few seeding nodes (sources) have non-zero values. Suppose further that as time evolves, nodes communicate with their neighbors according to some dynamics captured by $h_0,h_1,...$ , then the resultant state $\bbx$ can be represented as $\bbx= \sum_{l=0}^{\infty} h_l \bbS^l\bbz=\bbH\bbz$, i.e., the output of a graph filter to a sparse input graph signal $\bbz$. The expression $\bbx=\sum_{l=0}^{\infty} h_l \bbS^l\bbz$ with $\|\bbz\|_0\ll N$ has indeed been used to model a number of network dynamics as well as to solve different inverse problems involving observations of network processes \cite{segarra2017blind,djuric2018cooperative,segarra2017optimal,zhu2020estimating,zhu2020network}.

\vspace{.1cm}

\medskip\noindent\textbf{Stationary graph signals.} 
Consider now a statistical GSP setup where the values in $\bbx$ are random, and use $\bar{\bbx}=\E{\bbx}$ and $\bbC=\E{(\bbx-\bar{\bbx})(\bbx-\bar{\bbx})^\top}$ to denote the mean and covariance of this random process. 
In this setup, the \textit{random} graph process $\bbx$ is said to be stationary in the GSO $\bbS$ if its covariance matrix $\bbC$ is diagonalized by $\bbV$, the eigenvectors of the shift~\cite{marques2017stationary, perraudin2017stationary, girault2015translation}. 
Equivalently, a \textit{random} graph process is defined to be stationary in $\bbS$ if it can be represented as the output generated after filtering a white input with a linear graph filter $\bbH=\sum_{l=0}^{\infty} h_l \bbS^l$. 
Note that, when particularized to time-varying signals, the two aforementioned definitions boil down to the classical definition of stationary in time. 
The first definition requires stationary time processes to be uncorrelated in the Fourier domain, while the second one puts forth a generative model stating that a stationary time process can be represented as the output of a linear time-invariant filter to a white input~\cite{marques2017stationary}. 
More importantly for the graph context, the second definition reveals that covariance matrices of graph-stationarity signals can be written as (positive-semidefinite) polynomials of the GSO. In other words, the set of processes that are stationary on a (sparse) GSO $\bbS$ is formed by the random processes whose covariances can be written as polynomials of $\bbS$~\cite{marques2017stationary,segarra2017network}.
%%%%%%%%%%%%%%%%

%%%%%%%%%%%%%%%%%%%%%%%%%%%%%%%%%%%%%%%%%%%%%%%%%%%%%%%%%%%%%%
% PROBLEM STATEMENT
\section{Problem statement}\label{S:prelim_problem}

To state our joint network topology inference problem, start by considering a scenario with $K$ different graphs $\{\ccalG^{(k)}\}_{k=1}^K$ defined over the same set $\ccalN$ of nodes, but with possibly different sets of edges and weights. 
This implies that $K$ different GSOs $\{\bbS^{(k)}\}_{k=1}^K$ exist, each represented by an $N\times N$ matrix whose sparsity pattern and non-zero values may be different across $k$. 
Suppose also that, associated with each of the graphs, we have access to a set of graph signals collecting information attached to the nodes. 
Formally, we use matrix $\bbX^{(k)}:=[\bbx_1^{(k)},...,\bbx_{n_k}^{(k)}]\in\reals^{N\times n_k}$ to denote the matrix containing the $n_k$ graph signals associated with graph $\ccalG^{(k)}$. To simplify notation, we will assume that the signals are zero mean and denote the
\textit{sample} covariance of the $k$-th set as
\begin{equation}\label{E:sample_covariance_k}
\hat{\bbC}^{(k)} := \frac{1}{n_k}\bbX^{(k)}(\bbX^{(k)})^\top.
\end{equation}
The setup that we investigate in this paper is one where the \textit{graphs are unknown} and we want to use the \textit{observed signals to infer their topology}. 
This is feasible under the assumption that the properties of the signals are related to those of the underlying graph. 
Intuitively, when there is no relation among the different graphs, each of the $K$ topology inference problems can be solved separately.
However, if the graphs are related, joint inference can be beneficial. 
In this context, our problem is stated as follows.

%%%%%%%%%%%%%%%%%%%%%%%%%%%%%%%%%%
%% PROBLEM
\begin{problem}\label{problem_data_driven}
	Given the observations $\{\bbX^{(k)}\}_{k=1}^K$ find the graph structure encoded in $\{\bbS^{(k)}\}_{k=1}^K$ under the assumptions that: (AS1) the signals in $\bbX^{(k)}$ are realizations of a process that is stationary in $\bbS^{(k)}$ and (AS2) graphs $k$ and $k'$ are ``close'' according to a particular distance $d(\bbS^{(k)},\bbS^{(k')})$.
\end{problem}
%%%%%%%%%%%%%%%%%%%%%%%%%%%%%%%%%%

Although relatively formal, the statement of the problem above can give rise to different formulations. This issue will be resolved in Section \ref{S:ConvexRelax_RecovGuarantess},
%\ref{Ss:Shift_recov_via_nonlinear}
where an optimization problem associated with Problem 1 is presented. Before that, several remarks on assumptions (AS1) and (AS2) are provided. 
\vspace{.1cm}

%%%%%%%%%%%%%%%%
\medskip\noindent\textbf{(AS1) Stationarity:}
To better understand the implications of (AS1), let us recall that stationarity requires the covariance of the graph process to be a polynomial of $\bbS$. In other words, (AS1) is tantamount to assuming that the mapping between the GSO $\bbS$, which represents pairwise relationships between the nodes, and the matrix $\bbC=\E{\bbx\bbx^\top}$, which represents pairwise correlations between the nodes, is analytic (smooth), so that it can be accurately represented by a matrix polynomial. At an intuitive level, this model assumes that $\bbS$ encodes latent \textit{one-hop} interactions between nodes and that each successive application of the shift (i.e., higher-order powers of $\bbS$) spreads the original information across an iteratively increasing neighborhood, which ends up giving rise to \textit{indirect} correlations among all nodes in the graph \cite{djuric2018cooperative}. Put it differently, although the correlation is given by the \textit{dense} matrix $\bbC$, the actual dependencies can be (more easily) represented by the more \textit{parsimonious} matrix $\bbS$. Relevant relations between the shift and the covariance matrices that fall within this model include 
\begin{itemize}
	\item $\bbC=\bbS$, as in correlation networks; 
	\item $\bbC=\bbS^{-1}$, as in conditionally independent Markov random fields; or 
	\item $\bbC=(\bbI-\bbS)^{-2}$, as in symmetric structural equation models with white exogenous inputs.
\end{itemize} 
To elaborate on the third example, structural equation models postulate that the observed signal $\bbx$ can be written as $\bbx=\bbA \bbx + \bbw$, where $\bbw$ is the so-called exogenous input, and $\bbA$ is an adjacency matrix without self-loops \cite{shen2017kernel}. Rewriting the previous expression as $\bbx=(\bbI-\bbA)^{-1}\bbw$ and using the fact that $\bbw$ is white, it follows that $\E{\bbx\bbx^\top}=(\bbI-\bbA)^{T}\E{\bbw\bbw^\top}(\bbI-\bbA)^{-T}=(\bbI-\bbA)^{-2}$, where for the last step we have used that the graph is undirected. Note also that the second example, which can be equivalently written as $\bbS=\bbC^{-1}$, will allow us to establish meaningful links between our approach and graphical lasso. 
Although graph stationarity does not require Gaussianity, many of the works in the area assume that the graph signals at hand are not only stationary but also Gaussian distributed~\cite{djuric2018cooperative}. 
That is indeed the case for, e.g., linear network diffusion processes whose initial condition is Gaussian. While the algorithms presented in this paper can be applied regardless of the distribution of the data, the theoretical result in Theorem~\ref{T:robust_recovery} is the only point where Gaussianity is assumed.
%%%%%%%%%%%%%%%%

\vspace{.1cm}

%%%%%%%%%%%%%%%%
\medskip\noindent\textbf{(AS2) Similarity among graphs.}
Regarding (AS2), the two critical issues are the form of the distance function $d(\cdot,\cdot)$ and determining the proximity degree among the different graphs. 
To handle the second issue, let us define the weighted and directed graph $\ccalG_\ccalQ$ whose node set $\ccalQ$ collects the $K$ GSOs and with $W_{k,k'}$, the weight of edge $(k,k')$, representing the similarity between $\bbS^{(k)}$ and $\bbS^{(k')}$. 
The particular form of $\ccalG_\ccalQ$ will depend on the application at hand. 
In dynamic environments where the index $k$ corresponds to time, a reasonable choice is to set $\ccalG_\ccalQ$ to a \textit{directed path} connecting the GSOs corresponding to consecutive time instants (windows). 
Differently, if $k$ indexes patients with a particular disease, then it is reasonable to set $\ccalG_\ccalQ$ as a \textit{complete graph} with the strength of the connection $W_{k,k'}$ depending on the similarity between the corresponding patients. 
The weights in $\ccalG_\ccalQ$ can be known beforehand or learned from the data after postulating a particular model (see, e.g., \cite{oselio_hero2014multilayer} for a hierarchical approach). 
Regarding the form of $d(\bbS^{(k)},\bbS^{(k')})$, reasonable choices include $\|\vect(\bbS^{(k)}-\bbS^{(k')})\|_0$ and $\|\vect(\bbS^{(k)}-\bbS^{(k')})\|_1$, which will promote the pair of shifts to have the same sparsity pattern and weights; and $\|\vect(\bbS^{(k)}-\bbS^{(k')})\|_2^2$, which will promote similar weights.
Several of these distances have been explored in the context of joint identification of multiple sparse precision matrices $\bbC=\bbS^{-1}$ giving rise to modified graphical lasso formulations using regularized lasso %$\|\vect(\bbS^{(k)}-\bbS^{(k')})\|_1$ 
\cite{danaher2014joint}, regularized elastic net 
%$\alpha \|\vect(\bbS^{(k)}-\bbS^{(k')})\|_1+(1-\alpha)\|\vect(\bbS^{(k)}-\bbS^{(k')})\|_2$  
\cite{ryali2012estimation}, and regularized $\ell_{1,\infty}$ group lasso  
%$\|\bbS^{(k)}-\bbS^{(k')}\|_{1,\infty}$ 
\cite{honorio2010multi}. 
%%%%%%%%%%%%%%%%

%%%%%%%%%%%%%%%%%%%%%%%%%%%%%%%%%%%%%%%%%%%%%%%%%%%%%%%%%%%%%%
% CONVEX SOLUTION AND RECOVERY GUARANTEES
\section{Convex solution and recovery guarantees}\label{S:ConvexRelax_RecovGuarantess}

%%%%%%%%%%%%%%%%
Our goal is to provide an optimization-based solution to Problem 1. 
Specifically, our approach is to find the sparsest graphs $\{{\bbS^{(k)}}^*\}_{k=1}^K$ that satisfy assumptions (AS1) and (AS2) by solving 
\begin{alignat}{2}\label{eqn_zero_norm} 
\!\!&\!\min_{\{\bbS^{(k)}\}_{k=1}^K} \
&&\sum_{k} \alpha_k \|\mathrm{vec}(\bbS^{(k)})\|_0\;+\sum_{k < k'}  \beta_{k,k'}\, d(\bbS^{(k)},\bbS^{(k')})                                     \nonumber\\ 
\!\!&\!\mathrm{\;\;s. \;t. } && \!\!\!\!\!\bbC^{(k)}\bbS^{(k)} = \bbS^{(k)}\bbC^{(k)},  \;\;
\bbS^{(k)} \in \ccalS^{(k)}, \;\; \forall \;k. 
\end{alignat} 
In~\eqref{eqn_zero_norm}, the set $\ccalS^{(k)}$ specifies additional properties that $\bbS^{(k)}$ must satisfy, with examples including symmetry, zero diagonal elements (if the GSO is an adjacency matrix with no self-loops), or non-positive off-diagonal elements and $\bbzero=\bbS^{(k)}\bbone$ for the case of a combinatorial Laplacian. 
Regarding the structure of the objective, the first term promotes sparsity on the GSOs, the second one promotes the proximity postulated in (AS2), and $\{\alpha_k\}$ and $\{\beta_{k,k'}\}$ are parameters that allow a trade-off between the two terms in the objective. 
%Regarding the structure of the objective, the first term promotes sparsity on the GSOs, the second one promotes the proximity postulated in (AS2) and $\lambda$ is a parameter that allows to trade-off the two terms in the objective. 
Finally, the constraints $\bbC^{(k)}\bbS^{(k)} = \bbS^{(k)}\bbC^{(k)}$ account for (AS1). 
Specifically, note that stationarity implies that the eigenvectors of the covariance and those of the GSO are the same; hence, the covariance and the shift must commute, as enforced in the constraint. 
As will be apparent in Section~\ref{S:robust_recovery}, to take into account that in practice we have access to the \textit{sample} covariance $\hat{\bbC}^{(k)}$, it is reasonable to relax the equalities in $\bbC^{(k)}\bbS^{(k)} = \bbS^{(k)}\bbC^{(k)}$, with the level of tolerated violation depending on the number of samples available to form the estimates $\hat{\bbC}^{(k)}$.

\subsection{Relaxation for the sparse formulation}\label{Ss:relaxation}

Consider the following convex optimization problem
\begin{alignat}{2}\label{eqn_one_norm} 
\!\!&\!\min_{\{\bbS^{(k)}\}_{k=1}^K} \
&&\!\!\!\sum_{k} \alpha_k \|\mathrm{vec}(\bbS^{(k)})\|_1\!+\!\!\sum_{k < k'}  \beta_{k,k'}\, \| \mathrm{vec}(\bbS^{(k)} - \bbS^{(k')}) \|_1                \nonumber\\ 
\!\!&\!\mathrm{\;\;s. \;t. } && \bbC^{(k)}\bbS^{(k)} = \bbS^{(k)}\bbC^{(k)},  \;\;
\bbS^{(k)} = {\bbS^{(k)}}^\top, \;\; \forall \;k \nonumber \\
 & && S^{(k)}_{ii} = 0, \, \forall \; \{k,i\}, \,\,\, \textstyle\sum_{j=1}^{N} S^{(1)}_{j1} = 1.
\end{alignat} 

Notice that \eqref{eqn_one_norm} is a relaxed version of \eqref{eqn_zero_norm} where the $\ell_0$-norm has been replaced by the $\ell_1$-norm. 
Moreover, the distance $d(\cdot, \cdot)$ between the graph shifts was specialized to the $\ell_1$-norm of their difference and the feasibility sets $\ccalS^{(k)}$ were selected to represent symmetric adjacency matrices with zeros in the diagonal. 
Finally, the last constraint in \eqref{eqn_one_norm} fixes the scale of the recovered graphs and precludes the all-zero solution from belonging to the feasibility set. If we denote by $\{\hat{\bbS}^{(k)}\}_{k=1}^K$ the solution to \eqref{eqn_one_norm}, we now present conditions under which $\{\hat{\bbS}^{(k)}\}_{k=1}^K$ is guaranteed to coincide with the corresponding solution $\{{\bbS^{(k)}}^*\}_{k=1}^K$ to \eqref{eqn_zero_norm}. 

In order to formally define these conditions, a series of definitions must be put in place. 
First, define matrices $\bbB^{(i,j)} \in \reals^{N \times N}$ for $i < j$ such that $B^{(i,j)}_{ij} = 1$, $B^{(i,j)}_{ji} = -1$, and all other entries are zero. Based on this, we denote by $\bbB \in \reals^{{N \choose 2} \times N^2}$ a matrix whose rows are the vectorized forms of $\bbB^{(i,j)}$ for all $i,j \in \{1, 2, \ldots, N\}$ where $i < j$. In this way, $\bbB \bbs^{(k)} = \bbzero$ when $\bbs^{(k)}$ is the vectorized form of a symmetric matrix. Similarly, define vectors $\bbz^{(i,j)} \in \reals^{K}$ for $i < j \leq K$ such that $z^{(i,j)}_{i} = 1$, $z^{(i,j)}_{j} = -1$, and all other entries are zero. We build the matrix $\bbZ \in \reals^{ {K \choose 2} \times K}$ whose rows are the vectors ${\bbz^{(i,j)}}^\top$. 
We consolidate the information of all the covariances $\bbC^{(k)}$ in the block diagonal matrix $\bbSigma$ defined as $\bbSigma := \mathrm{blockdiag}(- \bbC^{(1)} \oplus \bbC^{(1)}, \ldots, - \bbC^{(K)} \oplus \bbC^{(K)})$ where, we recall, $\oplus$ denotes the Kronecker sum. 
With $\bbalpha$ and $\bbbeta$ collecting the values of \{$\alpha_k$\} and \{$\beta_{k,k'}$\} respectively, and $\ccalD' \!= \!\{1, N+2, \ldots , N^2\}$ denoting the indices corresponding to the diagonal of an $N \! \times \! N$ matrix when vectorized, we define the following two matrices
\begin{equation}\label{E:def_Psi_and_Phi}
	\bbPsi \!:=\! 
	\begin{bmatrix}
		\diag(\bbalpha) \\
		\diag(\bbbeta) \bbZ
	\end{bmatrix} \otimes \bbI_{N^2}, \quad
	\bbPhi \!:=\! 
	\begin{bmatrix}
		\bbI_K \otimes \bbB \\
		\bbI_K \otimes [\bbI_{N^2}]_{\ccalD'} \\
		\bbSigma \\
		(\bbe_1 \otimes \mathbf{1}_N)^\top
	\end{bmatrix}.
\end{equation}
Denote by $\ccalJ$ the index set of the support of $\bbs^*$, where $\bbs^* \in \reals ^{K N^2}$ collects the vectorized versions of $\{{\bbS^{(k)}}^*\}_{k=1}^K$, and by $\ccalI$ the index set of the support of $\bbPsi \bbs^*$. With this notation in place, the following result holds.

%%%%%%%%%%%%%%%%%%%%%%%%%%%%%%%%%%
%% PERFECT RECOVERY THEOREM
\begin{mytheorem}\label{T:theo_perfect_recovery}
Assuming problem \eqref{eqn_one_norm} is feasible, $\{\hat{\bbS}^{(k)}\}_{k=1}^K = \{{\bbS^{(k)}}^*\}_{k=1}^K$ if the two following conditions are satisfied:\\
	1) $[\bbPhi^\top]_{\ccalJ}$ is full row rank; and \\   
	2) {There exists a constant $\delta > 0$ such that 
	\begin{equation}\label{E:condition_recovery_noiseless}
		{\gamma} := \| \bbPsi_{\ccalI^c}(\delta^{-2}\bbPhi^\top \bbPhi + \bbPsi_{\ccalI^c}^\top \bbPsi_{\ccalI^c})^{-1}\bbPsi_{\ccalI}^\top \|_{\infty}<1.
% 		{\gamma} := \| \bbPsi_{\ccalI^c}(\delta^{-2}\bbPhi^\top \bbPhi + \bbPsi_{\ccalI^c}^\top \bbPsi_{\ccalI^c})^{-1}\bbPsi_{\ccalI}^\top \|_{\infty}<1.
	\end{equation}}
\end{mytheorem}
\begin{myproof}
Denoting by $\bbs^{(k)} = \mathrm{vec}(\bbS^{(k)})$ for all $k$, problem~\eqref{eqn_one_norm} can be reformulated as
\begin{alignat}{2}\label{eqn_one_norm_vectorized} 
\!\!&\!\min_{\{\bbs^{(k)}\}_{k=1}^K} \
&& \sum_{k} \alpha_k \|\bbs^{(k)}\|_1\;+\sum_{k < k'}  \beta_{k,k'}\, \| \bbs^{(k)} - \bbs^{(k')} \|_1                \nonumber\\ 
\!\!&\!\mathrm{\;\;s. \;t. } && (\bbI_{N} \otimes \bbC^{(k)} \!-\! \bbC^{(k)} \otimes \bbI_{N}) \bbs^{(k)} \! = \mathbf{0},  \;  \nonumber \\ 
& && \bbB \bbs^{(k)} \!= \! \mathbf{0}, \,\,\,\,   [\bbI_{N^2}]_{\ccalD'} \,\, \bbs^{(k)} = \mathbf{0}, \quad\,\,\,\,\,  \forall \; k \nonumber \\
& && (\bbe_1 \otimes \mathbf{1}_N)^\top \bbs^{(1)} = 1,
\end{alignat} 
where, we recall, $\bbs^{(k)}$ belonging to the null space of $\bbB$ ensures that $\bbS^{(k)}$ is symmetric, and the last equality imposes that the first column of $\bbS^{(1)}$ sums up to 1 [cf. last constraint in~\eqref{eqn_one_norm}].
Denoting by $\bbs = [{\bbs^{(1)}}^\top, \ldots, {\bbs^{(K)}}^\top]^\top$ and leveraging the definitions in \eqref{E:def_Psi_and_Phi}, problem \eqref{eqn_one_norm_vectorized} can be compactly stated as
\begin{equation}\label{eqn_one_norm_vectorized_compact} 
\min_{\bbs} \, \| \bbPsi \bbs \|_1      \quad \mathrm{\;\;s. \;t. } \,\, \bbPhi \bbs = \bbb,
\end{equation} 
where $\bbb$ is a binary vector of length $K({N \choose 2}+ N^2+N) +1$ with all its entries equal to $0$ except for the last one that is a $1$. Problem~\eqref{eqn_one_norm_vectorized_compact} is an instance of $\ell_1$-analysis \cite{zhang2016one}. It can be shown \cite[Theorem 1] {zhang2016one} that the solution to \eqref{eqn_one_norm_vectorized_compact} coincides with the sparsest solution $\bbs^*$ if:

\begin{itemize}
	\item[a)] $\mathrm{ker}(\bbPsi_{\ccalI^c}) \cap \mathrm{ker}(\bbPhi) = \{ \mathbf{0} \}$; and
	\item[b)] There exists a vector $\bby \in \reals^{N^2 (K + {K \choose 2})}$ such that $\bbPsi^\top \bby \in \mathrm{Im}(\bbPhi^\top)$, $\bby_\ccalI = \mathrm{sign}(\bbPsi_{\ccalI} \bbs^*)$, and $\|\bby_{\ccalI^c}\|_\infty < 1$.
\end{itemize}
\noindent The remainder of the proof is devoted to showing that if conditions \emph{1)} and \emph{2)} in the statement of the theorem hold, then a) and b) are satisfied.

We begin by showing that \emph{1)} implies a). In order to do this, we first provide some insight on the specific form of $\bbPsi_{\ccalI^c}$. 
Notice that the first $K \, N^2$ rows of $\bbPsi$ correspond to the computation of the $\ell_1$-norm cost of each entry of the $K$ graph shifts whereas the last ${K \choose 2} \, N^2$ rows of $\bbPsi$ correspond to the cost of a discrepancy between corresponding entries of two different graph shifts. Hence, the rows selected in $\bbPsi_{\ccalI^c}$, i.e., the ones \emph{not} in the support of $\bbPsi \, \bbs^*$, belong to two classes: i) among the first $K \, N^2$ rows, $\ccalI^c$ selects the rows corresponding to elements in $\bbs^*$ which are $0$; and ii) among the last ${K \choose 2} \, N^2$, $\ccalI^c$ selects the rows corresponding to pairs of elements that are repeated in two different graph shifts. Thus, for a generic vector $\bbw \in \reals^{K \, N^2}$ to belong to $\mathrm{ker}(\bbPsi_{\ccalI^c})$ two conditions must be satisfied (associated with the two aforementioned classes): i) if $s^*_i = 0$ then $w_i = 0$; and ii) if $s^*_{(k-1) N^2 + i} = s^*_{(k'-1) N^2 + i}$ for some $k, k', i$ then $w_{(k-1) N^2 + i} = w_{(k'-1) N^2 + i}$. For a) to be satisfied, we need to guarantee that any such $\bbw$ cannot belong to the null space of $\bbPhi$. A sufficient condition for this is to require that columns $i$ of $\bbPhi$ associated with values $\bbs_i^* \neq 0$ are linearly independent, which is exactly condition \emph{1)} in the theorem's statement.

The next step is to show that condition \emph{2)} implies b). For this, consider the following $\ell_2$ norm minimization problem

\begin{align}\label{E:l_2_minimization_dual_certificate}
\min_{ \{\bby, \bbr\}}  \delta^2\| \bbr \|_2^2 + \|\bby\|_2^2  \;\;\text{{s. t.} } \; \bbPsi^\top \bby= \bbPhi^\top \bbr,\;\; \bby_\ccalI =  \mathrm{sign}(\bbPsi_{\ccalI} \bbs^*), 
\end{align}
where $\delta$ is a positive tuning constant. Including the term $\delta^2 \|\bbr\|_2^2$ in the objective guarantees the existence of a closed-form expression for the minimizing argument, while preventing numerical instability when solving the optimization. We now show that the solution $\bby^*$ of \eqref{E:l_2_minimization_dual_certificate} satisfies the requirements imposed in condition b). The two constraints in \eqref{E:l_2_minimization_dual_certificate} enforce the fulfillment of the first two requirements in b), hence, we are left to show that $\| \bby^*_{\ccalI^c}\|_\infty < 1$.
Since the values of $\bby_{\ccalI}$ are fixed, the constraint $\bbPsi^\top \bby= \bbPhi^\top \bbr$ can be rewritten as $\bbPsi_{\ccalI}^\top \mathrm{sign}(\bbPsi_{\ccalI} \bbs^*) = -\bbPsi_{\ccalI^c}^\top \bby_{\ccalI^c} +  \bbPhi^\top \delta^{-1} \delta\bbr$. Then, by defining the vector $\bbt := [\delta\bbr^\top, -\bby_{\ccalI^c}^\top]^\top$ and the matrix $\bbQ:= [\delta^{-1} \bbPhi^\top, \bbPsi_{\ccalI^c}^\top]$, \eqref{E:l_2_minimization_dual_certificate} can be rewritten as
\begin{align}\label{E:l_2_minimization_dual_certificate_rewritten}
\min_{\bbt} \;\; \|\bbt\|_2^2  \;\;\text{{s. t.} } \; \bbPsi_{\ccalI}^\top \mathrm{sign}(\bbPsi_{\ccalI} \bbs^*)  = \bbQ \bbt.
\end{align}
The minimum-norm solution to \eqref{E:l_2_minimization_dual_certificate_rewritten} is given by $\bbt^*=(\bbQ)^\dag \bbPsi_{\ccalI}^\top \mathrm{sign}(\bbPsi_{\ccalI} \bbs^*)$ from where it follows that
\begin{align}\label{E:dual_certificate_solution_norm2_delta}
\bby^*_{\ccalI^c}  \!=\!  - \bbPsi_{\ccalI^c}(\delta^{-2}\bbPhi^\top \bbPhi+\bbPsi_{\ccalI^c}^\top \bbPsi_{\ccalI^c})^{-1}\bbPsi_{\ccalI}^\top \, \mathrm{sign}(\bbPsi_{\ccalI} \bbs^*).
\end{align}
%
% Condition a) guarantees the existence of the inverse in \eqref{E:dual_certificate_solution_norm2_delta}. Since $\| \mathrm{sign}(\bbPsi_{\ccalI} \bbs^*) \|_{\infty} \! = \! 1$, we may bound the $\ell_\infty$ norm of $\bby^*_{\ccalI^c}$ as $\| \bby^*_{\ccalI^c} \|_{\infty} \leq \| \bbPsi_{\ccalI^c}(\delta^{-2}\bbPhi^\top \bbPhi+\bbPsi_{\ccalI^c}^\top \bbPsi_{\ccalI^c})^{-1}\bbPsi_{\ccalI}^\top \|_{M(\infty)} = \gamma$.
Condition a) guarantees the existence of the inverse in \eqref{E:dual_certificate_solution_norm2_delta}. Since $\| \mathrm{sign}(\bbPsi_{\ccalI} \bbs^*) \|_{\infty} \! = \! 1$, we may bound the $\ell_\infty$ norm of $\bby^*_{\ccalI^c}$ as $\| \bby^*_{\ccalI^c} \|_{\infty} \leq \| \bbPsi_{\ccalI^c}(\delta^{-2}\bbPhi^\top \bbPhi+\bbPsi_{\ccalI^c}^\top \bbPsi_{\ccalI^c})^{-1}\bbPsi_{\ccalI}^\top \|_{\infty} = \gamma$.
Hence, condition \emph{2)} in the theorem guarantees $\| \bby^*_{\ccalI^c} \|_{\infty}<1$ as wanted.
\end{myproof}
%%%%%%%%%%%%%%%%%%%%%%%%%%%%%%%%%%

Theorem~\ref{T:theo_perfect_recovery} provides \emph{sufficient} conditions under which the relaxation in~\eqref{eqn_one_norm}  is guaranteed to recover the true sparse GSOs $\{{\bbS^{(k)}}^*\}_{k=1}^K$.
Numerical experiments in Section~\ref{S:numerical_experiments} reveal that the bound imposed on $\gamma$ in~\eqref{E:condition_recovery_noiseless} is tight by providing examples where $\gamma=1$ and for which recovery fails. 
In the statement of the theorem, condition \emph{1)} ensures that the solution to~\eqref{eqn_one_norm} is unique, a necessary requirement to guarantee sparse recovery. 
Condition \emph{2)} is derived from the construction of a dual certificate designed to ensure that the unique solution to~\eqref{eqn_one_norm} also has minimum $\ell_0$ (pseudo-)norm~\cite{zhang2016one}.
Details within the proof of the theorem reveal why condition \emph{2)} is sufficient but not necessary.
In a nutshell, the condition guarantees that a \emph{specific} judicious candidate for the dual certificate (obtained by minimizing a relevant $\ell_2$ norm) satisfies a bound on its $\ell_\infty$ norm. 
However, when this specific candidate fails, one cannot rule out the existence of better dual certificates that can ensure sparse recovery.
To gain further intuition on~\eqref{E:condition_recovery_noiseless}, notice that condition \emph{2)} is always satisfied whenever $\bbPhi^\top \bbPhi$ is invertible. 
Indeed, for small values of $\delta$ we have that $\gamma \approx \delta^2 \| \bbPsi_{\ccalI^c}(\bbPhi^\top \bbPhi)^{-1}\bbPsi_{\ccalI}^\top \|_{\infty}$, which can be made smaller than $1$ by selecting arbitrarily small values of $\delta$.
This should not be surprising since $\bbPhi^\top \bbPhi$ being invertible implies that $\bbPhi$ has full column rank which, in turn, implies that the feasibility set of our problem is a singleton [cf.~\eqref{eqn_one_norm_vectorized_compact}].
Thus, in this extreme case, the $\ell_1$ relaxation (and any other objective) is guaranteed to recover the true GSOs.
Notice that the guarantees for exact recovery provided by Theorem~\ref{T:theo_perfect_recovery} strongly rely on the fact that all constraints in~\eqref{eqn_one_norm} are equality constraints. 
This, in turn, is enabled by the assumption that we have perfect knowledge of the covariances $\bbC^{(k)}$.
Thus, the more practical scenario where the covariances are estimated requires a robust reformulation of the recovery problem, as we discuss next.

%%%%%%%%%%%%%%%%%%%%%%%%%%%%%%%%%%%%%%%%%%%%%%%%%%%%%%%%%%%%%%
% ROBUST RECOVERY
\section{Robust recovery and sample complexity}\label{S:robust_recovery}

Following the formal description of Problem~\ref{problem_data_driven}, we do not have access to the covariance matrices $\bbC^{(k)}$ but rather to signals $\{\bbX^{(k)}\}_{k=1}^K$.
Hence, we reformulate~\eqref{eqn_one_norm} to account for the fact that we can only have access to sample estimates $\hat{\bbC}^{(k)}$ of the covariances [cf.~\eqref{E:sample_covariance_k}].
More specifically, the commutativity constraint in \eqref{eqn_one_norm}, $\bbC^{(k)}\bbS^{(k)} = \bbS^{(k)}\bbC^{(k)}$, is relaxed and instead we bound the difference between $\hat{\bbC}^{(k)} \bbS^{(k)}$ and $\bbS^{(k)} \hat{\bbC}^{(k)}$, giving rise to the following optimization problem
\begin{alignat}{2}\label{E:l1_norm} 
\!\!&\!\min_{\{\bbS^{(k)}\}_{k=1}^K} \
&&\!\!\sum_{k} \alpha_k \|\mathrm{vec}(\bbS^{(k)})\|_1\!+\!\sum_{k < k'}  \beta_{k,k'}\, \| \mathrm{vec}(\bbS^{(k)} - \bbS^{(k')}) \|_1                \nonumber\\ 
\!\!&\!\mathrm{\;\;s. \;t. } && \sum_{k=1}^K \| \bbS^{(k)} \hat{\bbC}^{(k)} - \hat{\bbC}^{(k)} \bbS^{(k)} \|_{\mathrm{F}}^2 \leq \epsilon_n^2 \nonumber\\
%\bbC^{(k)}\bbS^{(k)} = \bbS^{(k)}\bbC^{(k)},  \;\;
 & && \bbS^{(k)} = {\bbS^{(k)}}^\top, \;\; \forall \;k \nonumber \\
 & && S^{(k)}_{ii} = 0, \, \forall \; \{k,i\}, \,\,\, \textstyle\sum_{j=1}^{N} S^{(1)}_{j1} = 1.
\end{alignat} 

Our goal is to bound the distortion between the real GSOs $\{ {\bbS^{(k)}}^* \}_{k=1}^K$ and the estimated ones $\{ \hat{\bbS}^{(k)} \}_{k=1}^K$ obtained by solving~\eqref{E:l1_norm}, where $\epsilon_n$ is selected large enough to ensure feasibility.
To formally state this bound, a series of definitions must be put in place.

Recalling that $\bbs^* \in \reals^{KN^2}$ collects the vectorized versions of the true GSOs, $\{{\bbS^{(k)}}^*\}_{k=1}^K$, we denote by $\ccalD$, $\ccalL$, and $\ccalU$ the indices in $\bbs^*$ corresponding to the diagonal, lower triangular, and upper triangular elements of ${\bbS^{(k)}}^*$ for $k = 1, \ldots, K$. 
%More specifically, $\bbs_\ccalD = [\diag(\bbS^{(1)})^\top, \ldots, \diag(\bbS^{(K)})^\top]^\top$, $\bbs_{\ccalL} = [\mathrm{tril}(\bbS^{(1)})^\top, \ldots, \mathrm{tril}(\bbS^{(K)})^\top]^\top$, and $\bbs_{\ccalU} = [\mathrm{tril}({\bbS^{(1)}}^\top)^\top, \ldots, \mathrm{tril}({\bbS^{(K)}}^\top)^\top]^\top$.
Analogous to the definition of $\bbSigma$, we define the block diagonal $\hat{\bbSigma}$ that combines the sample covariance matrices $\hat{\bbC}^{(k)}$ as $\hat{\bbSigma} := \mathrm{blockdiag}(- \hat{\bbC}^{(1)} \oplus \hat{\bbC}^{(1)}, \ldots, - \hat{\bbC}^{(K)} \oplus \hat{\bbC}^{(K)})$. 
Define matrices $\bbM := [\hat{\bbSigma}]^\top_{\ccalL} + [\hat{\bbSigma}]^\top_{\ccalU} \in \reals^{K N^2 \times K \binom{N}{2}}$, $\bbR := [\bbPsi^\top]_\ccalL^\top \in \reals^{\left(K + \binom{K}{2}\right) N^2 \times K \binom{N}{2}}$, and let $\sum_k n_k = n$. We let $\ccalK$ represent the support of $\bbR \bbs_\ccalL^*$. Finally, we define the constant $\omega := \max_{k=1,\dots,K} \omega_k$ where $\omega_k := \max \{ \max_i [\bbC^{(k)}]_{ii}, \max_i [{\bbS^{(k)}}^* \bbC^{(k)} {\bbS^{(k)}}^*]_{ii} \}$.
With this notation in place, we state our main result on the performance of the proposed robust recovery scheme.

%%%%%%%%%%%%%%%%%%%%%%
% ROBUST RECOVERY THEOREM
\begin{mytheorem}\label{T:robust_recovery} 
If the following five conditions are satisfied:\\
1) $\bbM$ is full column rank. \\
2) $K = o(\log N )$.\\
3) $n_1 \asymp n_2 \asymp \ldots \asymp n_K$. \\
4) $\log N = o(\min\{n/(K^7 (\log n)^2), (n/K^7)^{1/3}\})$.\\
5) $\epsilon_n \geq C N \omega \sqrt{(K \log N)/n} $, for some constant $C > 0$. \\
Then, with probability at least $1 - e^{- C' \log N}$ for some constant $C'>0$ we have that 
\begin{equation}\label{E:theorem_robust}
\begin{aligned}
&\sum_{k=1}^K \| \mathrm{vec}(\hat{\bbS}^{(k)} - {\bbS^{(k)}}^*) \|_1 \leq \gamma \epsilon_n, \\
&\text{where} \,\,\, \gamma = \frac{4 \sqrt{|\ccalK|}\sigma_{\max}(\bbR) \| \bbR^\dag \|_{1}}{\sigma_{\min}(\bbM)} \left(2+\sqrt{|\ccalK|}\right).
\end{aligned}
\end{equation}
\end{mytheorem}
\begin{myproof}
We first state the following lemma, which characterizes the eigenvalues of a matrix after performing a rank-one update and that will be instrumental in showing our main result.

%%%%%%%%%%%
% LEMMA 3
\begin{mylemma}\cite{Golub73}
	\label{L:matrix_eigenvalues}
	Let $\bbC = \bbD + \bbu \bbu^\top$ where $\bbD = \diag(\bbd)$ is a diagonal matrix of size $m \times m$ such that $d_i \leq d_{i+1}$. We denote the eigenvalues of $\bbC$ by $\lambda_i$ such that $\lambda_i \leq \lambda_{i+1}$. Then, for $i = 1, \ldots, m-1$ it holds that $d_i \leq \lambda_i \leq d_{i+1}$.
\end{mylemma}
%
%%%%%%%%%%%

Recalling that $\bbs = [\vect(\bbS^{(1)})^\top, \dots, \vect(\bbS^{(K)})^\top]^\top$, we may reformulate~\eqref{E:l1_norm} as
\begin{align}\label{E:proof_recovery_010}
	\!\!&\! \min_{\bbs} && \| (\diag(\bbalpha) \otimes \bbI_{N^2}) \, \bbs\|_1 + \| (\diag(\bbbeta) \bbZ \, \otimes \, \bbI_{N^2}) \, \bbs\|_1 \nonumber \\
	\!\!&\!\mathrm{\;\;s. \;t. }  && \| \hat{\bbSigma} \, \bbs \|_2 \leq \epsilon_n, \,\,\,\,  \bbs_\ccalD = \mathbf{0}, \,\,\,\, \bbs_{\ccalL} = \bbs_{\ccalU}, \,\,\,\, \nonumber\\
%	& && (\bbI_K \otimes \bbe_1 \otimes \bbone_p)^\top \bbs = \mathbf{1},
	& && (\bbe_1 \otimes \bbone_N)^\top \bbs = 1,
\end{align}
where the second, third, and fourth constraints correspond to the feasibility conditions in \eqref{E:l1_norm}. 
Decomposing $\bbs$ into $\bbs_\ccalD$, $\bbs_{\ccalL}$, and $\bbs_{\ccalU}$, we may write the first constraint in \eqref{E:proof_recovery_010} as $\| [\hat{\bbSigma}]^\top_{\ccalD} \, \bbs_{\ccalD} + [\hat{\bbSigma}]^\top_{\ccalL} \, \bbs_{\ccalL} + [\hat{\bbSigma}]^\top_{\ccalU} \, \bbs_{\ccalU} \|_2 \leq \epsilon_n$. This enables us to restate \eqref{E:proof_recovery_010} only in terms of $\bbs_{\ccalL}$ as follows
\begin{equation}\label{E:proof_recovery_020}
\begin{aligned}
	& \hat{\bbs}_{\ccalL} =  && \argmin_{\bbs_{\ccalL}} \,\,\,  \|\bbR \bbs_{\ccalL}\|_1 \quad\quad \\
	\!\!&\!\mathrm{\;\;s. \;t. }  \,\,\,\, && \| \bbM  \, \bbs_{\ccalL} \|_2 \leq \epsilon_n, \,\,\,\, (\bbe_1 \otimes \bbone_{N-1})^\top \bbs_{\ccalL} = 1,
%	\!\!&\! \text{{s. to} }  \,\,\,\, && \| \bbM  \, \bbs_{\ccalL} \|_2 \leq \epsilon_n, \,\,\,\, (\bbe_1 \otimes \bbone_{N-1})^\top \bbs_{\ccalL} = \mathbf{1},
%	\!\!&\! \text{{s. to} }  \,\,\,\, && \| \bbM  \, \bbs_{\ccalL} \|_2 \leq \epsilon_n, \,\,\,\, (\bbI_K \otimes \bbe_1 \otimes \bbone_{N-1})^\top \bbs_{\ccalL} = \mathbf{1}.
\end{aligned}
\end{equation}
where we have assumed that $N$ is even to simplify the notation in the last constraint in~\eqref{E:proof_recovery_020}.
We now introduce a slight variation on problem \eqref{E:proof_recovery_020} parametrized by $q > 0$, where we relax the equality constraint as follows
\begin{equation}\label{E:proof_recovery_030}
\begin{aligned}
& \hat{\bbs}_{\ccalL}^{(q)} =  && \argmin_{\bbs_{\ccalL}} \,\,\,  \| \bbR \bbs_{\ccalL}\|_1 \quad\quad \\
\!\!&\!\mathrm{\;\;s. \;t. }   \,\,\,\, && \left\| \begin{bmatrix} \bbM \\ q\, (\bbe_1 \otimes \bbone_{N-1})^\top \end{bmatrix} \, \bbs_{\ccalL} - \begin{bmatrix} \mathbf{0} \\ q \end{bmatrix}\right\|_2 \leq \epsilon_n.
%\!\!&\! \text{{s. to} }  \quad \left\| \begin{bmatrix} \bbM \\ q\, (\bbe_1 \otimes \bbone_{N-1})^\top \end{bmatrix} \, \bbs_{\ccalL} - \begin{bmatrix} \mathbf{0} \\ q \mathbf{1} \end{bmatrix}\right\|_2 \leq \epsilon_n.
%\!\!&\! \min_{\bbs_{\ccalL}} \,\,\, \| \bbR \bbs_{\ccalL}\|_1 \quad\quad \\
%\!\!&\! \text{{s. to} }  \quad \left\| \begin{bmatrix} \bbM \\ q\, (\bbI_K \otimes \bbe_1 \otimes \bbone_{N-1})^\top \end{bmatrix} \, \bbs_{\ccalL} - \begin{bmatrix} \mathbf{0} \\ q \mathbf{1} \end{bmatrix}\right\|_2 \leq \epsilon_n.
\end{aligned}
\end{equation}

Notice that parameter $q$ controls the admissible level of violation of the original equality constraint in \eqref{E:proof_recovery_020}. In particular, for large $q$ the equality must hold, i.e., $\lim_{q \to \infty} \hat{\bbs}_{\ccalL}^{(q)} = \hat{\bbs}_{\ccalL}$. 
For notational convenience, let us define $ \bbt_q = q\, (\bbe_1 \otimes \bbone_{N-1})$, 
$\bbPhi_q = [ \bbM^\top , \, \bbt_q]^\top$, and $\bbb_q = [ \mathbf{0}^\top, \, q]^\top$, 
where we explicitly state their dependence on the parameter $q$. 
In Claim~\ref{C:problem_q} we prove recovery conditions for problem \eqref{E:proof_recovery_030}, where the parameter $q$ plays a central role. More precisely, we bound the distance between the solution $\hat{\bbs}_{\ccalL}^{(q)}$ for \eqref{E:proof_recovery_030} and the true graph $\bbs^*_{\ccalL}$. 
The proof of this claim is deferred to the appendix.

%%%%%%%%%%%%%%%%%%%%%%%%%%%%%%%%%%
%% CLAIM 1
\begin{myclaim}\label{C:problem_q}
If the following two conditions are satisfied:\\
l1) $\bbPhi_q$ is full column rank. \\
l2) $ \| \bbPhi_q \bbs^*_{\ccalL} - \bbb_q \|_2 \leq \epsilon_n $. \\
Then, we have that
\begin{equation}\label{E:def_c_q}
\begin{aligned}
&\| \hat{\bbs}_{\ccalL}^{(q)} - \bbs^*_{\ccalL} \|_1 \leq \gamma_q \epsilon_n, \qquad \\
&\text{where} \,\,\, \gamma_q = \frac{2 \sqrt{|\ccalK|}\sigma_{\max}(\bbR) \| \bbR^\dag \|_{1}}{\sigma_{\min}(\bbPhi_q)} \left(2+\sqrt{|\ccalK|}\right) .
\end{aligned}
\end{equation}
\end{myclaim}
%%%%%%%%%%%%%%%%%%%%%%%%%%%%%%%%%%

We now show that requirements \emph{1)}-\emph{5)} in the statement of Theorem~\ref{T:robust_recovery} imply conditions \emph{l1)} and \emph{l2)} in Claim~\ref{C:problem_q} as $q \to \infty$. 
That \emph{1)} implies \emph{l1)} follows from a simple argument. Indeed, given that $\bbPhi_q$ is generated from $\bbM$ by adding the row corresponding to 
$q\, (\bbe_1 \otimes \bbone_{N-1})^\top$, 
%$q\, (\bbI_K \otimes \bbe_1 \otimes \bbone_{N-1})^\top$, 
the column rank of $\bbPhi_q$ cannot be smaller than that of $\bbM$. Since $\bbM$ is full column rank, $\bbPhi_q$ must be as well.
That \emph{2)}-\emph{5)} imply \emph{l2)} is shown in the following claim, whose proof is also deferred to the appendix.

%%%%%%%%%%%%%%%%%%%%%%%%%%%%%%%%%%
%% CLAIM 2
\begin{myclaim}\label{C:reqs_3}
If conditions \emph{2)}-\emph{5)} in the statement of Theorem~\ref{T:robust_recovery} hold, then with probability at least $1 - e^{- C' \log N}$ for some constant $C'>0$ we have that $ \| \bbPhi_q \bbs^*_{\ccalL} -  \bbb_q \|_2 \leq \epsilon_n $ as $q \to \infty$.
\end{myclaim}
%%%%%%%%%%%%%%%%%%%%%%%%%%%%%%%%%%

Recall that the solution $\hat{\bbs}_{\ccalL}$ of problem \eqref{E:proof_recovery_020} coincides with $\hat{\bbs}_{\ccalL}^{(q)} $ for $q \to \infty$. Hence, from Claims~\ref{C:problem_q} and \ref{C:reqs_3}, it follows that under the conditions of Theorem~\ref{T:robust_recovery} it holds with high probability that $\| \hat{\bbs}_{\ccalL} - \bbs^*_{\ccalL} \|_1 \leq \gamma_\infty \epsilon_n$, where $\gamma_\infty := \lim_{q \to \infty} \gamma_q$. Moreover, in terms of the full matrices (instead of just the lower triangular components), this implies that 
\begin{equation}\label{E:full_matrices_bound}
\sum_{k=1}^K \| \mathrm{vec}(\hat{\bbS}^{(k)} - {\bbS^{(k)}}^*) \|_1 \leq 2 \, \gamma_\infty \epsilon_n.
\end{equation}
Consequently, if we show that $2 \, \gamma_\infty \leq \gamma$ as defined in~\eqref{E:theorem_robust} the proof concludes. More specifically, we want to show that
\begin{equation}\label{E:bound_limits}
\begin{aligned}
&\lim_{q \to \infty} \, \frac{4 \sqrt{|\ccalK|}\sigma_{\max}(\bbR) \| \bbR^\dag \|_{1} }{\sigma_{\min}(\bbPhi_q)} \left(2+\sqrt{|\ccalK|}\right) \,\, \\ 
&\qquad \leq  \,\, \frac{4 \sqrt{|\ccalK|}\sigma_{\max}(\bbR) \| \bbR^\dag \|_{1} }{\sigma_{\min}(\bbM)} \left(2+\sqrt{|\ccalK|}\right).
\end{aligned}
\end{equation}
This boils down to showing that $\lim_{q \to \infty} \sigma_{\min}(\bbPhi_q) \geq \sigma_{\min}(\bbM)$. 
To prove this, we use Lemma~\ref{L:matrix_eigenvalues}. 
Let us denote the eigendecomposition of the real and symmetric matrix $\bbM^\top \bbM = \bbV \bbD \bbV^\top$. 
Notice that the singular values of $\bbM$ are then given by $\sqrt{d_i}$. 
From the definition of $\bbPhi_q$ it readily follows that $\bbPhi_q^\top \bbPhi_q = \bbM^\top \bbM + \bbt_q \bbt_q^\top$. 
We may rewrite this equality in a form more amenable to Lemma~\ref{L:matrix_eigenvalues} as $\bbV^\top \bbPhi_q^\top \bbPhi_q \bbV = \bbD + \bbV^\top  \bbt_q \bbt_q^\top \bbV$. 
Equating $\bbV^\top \bbPhi_q^\top \bbPhi_q \bbV$ to $\bbC$ and $\bbV^\top  \bbt_q $ to $\bbu$ in Lemma~\ref{L:matrix_eigenvalues}, we obtain that $\sigma_{\min}(\bbPhi_q) \geq \sigma_{\min}(\bbM)$ for all $q > 0$. 
In particular, $\lim_{q \to \infty} \sigma_{\min}(\bbPhi_q) \geq \sigma_{\min}(\bbM)$ as we wanted to show, concluding the proof of the theorem.
\end{myproof}
%
%%%%%%%%%%%%%%%%%%%%%%

Theorem~\ref{T:robust_recovery} provides a high-probability bound on the error incurred when solving~\eqref{E:l1_norm}.
The fact that the result is probabilistic in nature is expected.
Indeed, since we are estimating the covariances from observed signals, there is always a small chance that the estimates are too noisy to enable approximate recovery of the GSOs.
Let us now analyze the five conditions required in the statement of the theorem.
Condition~\emph{1)} is akin to requiring the feasibility set to be a singleton when perfect covariances are available.
To see this, notice that if in~\eqref{E:proof_recovery_020} we make $\epsilon_n=0$ and $\bbM$ is full column rank, then $\bbs_\ccalL$ is completely determined.
Relating this back to Theorem~\ref{T:theo_perfect_recovery}, recovery in the noiseless case is guaranteed in this setting [cf. discussion after Theorem~\ref{T:theo_perfect_recovery}]. 
However, the current theorem describes how this recovery degrades with noise in the estimated covariances.
Condition~\emph{3)} imposes the reasonable requirement that the amount of signals observed from each graph is comparable. 
Intuitively, since our objective is to gain inference power by pooling signals together, the case where only a vanishing number of signals are associated with a specific graph is detrimental to the estimation of that graph.
Conditions~\emph{2)} and~\emph{4)} impose relations between the size of the graphs~$N$, the number of signals available~$n$, and the number of graphs~$K$. 
The number of graphs should be small in relation to the number of nodes in each of those graphs [cf. condition~\emph{2)}] and the number of nodes cannot be too large compared with the number of observed signals [cf. condition~\emph{4)}].
Condition~\emph{5)} provides a direct handle on the recovery error by determining the minimum admissible $\epsilon_n$.
More precisely, if $\epsilon_n$ is too small then the problem might become infeasible or no approximate solution might be included in the feasibility set. 
On the other hand, if $\epsilon_n$ is chosen too large then the bound in~\eqref{E:theorem_robust} would be too loose.
In this context, condition~\emph{5)} guides the choice of $\epsilon_n$ so that it is large enough for the result to hold while trying to minimize the upper bound on the estimation error.
Consistent with our discussion of condition~\emph{1)}, whenever $n \to \infty$, we have that $\epsilon_n \to 0$ and~\eqref{E:theorem_robust} guarantees a perfect recovery.
More interestingly, Theorem~\ref{T:robust_recovery} reveals the behavior of this error for finite values of $n$.
Indeed, for fixed $K$ and $N$, $\epsilon_n$ decreases as $1/\sqrt{n}$ and the only term in $\gamma$ dependent on $n$ is $\sigma_{\min}(\bbM)$. 
The revealed functional dependence arises in practice, as we illustrate in the next section.

%%%%%%%%%%%%%%%%%%%%%%%%%%%%%%%%%%%%%%%%%%%%%%%%%%%%%%%%%%%%%%
% NUMERICAL EXPERIMENTS
\section{Numerical experiments}\label{S:numerical_experiments}

%%%%%%%%%%%%%%%%%%%%%%
% FIGURE
\begin{figure*}
	\centering
	
	\begin{minipage}[c]{.25\textwidth}
		\includegraphics[width=\textwidth]{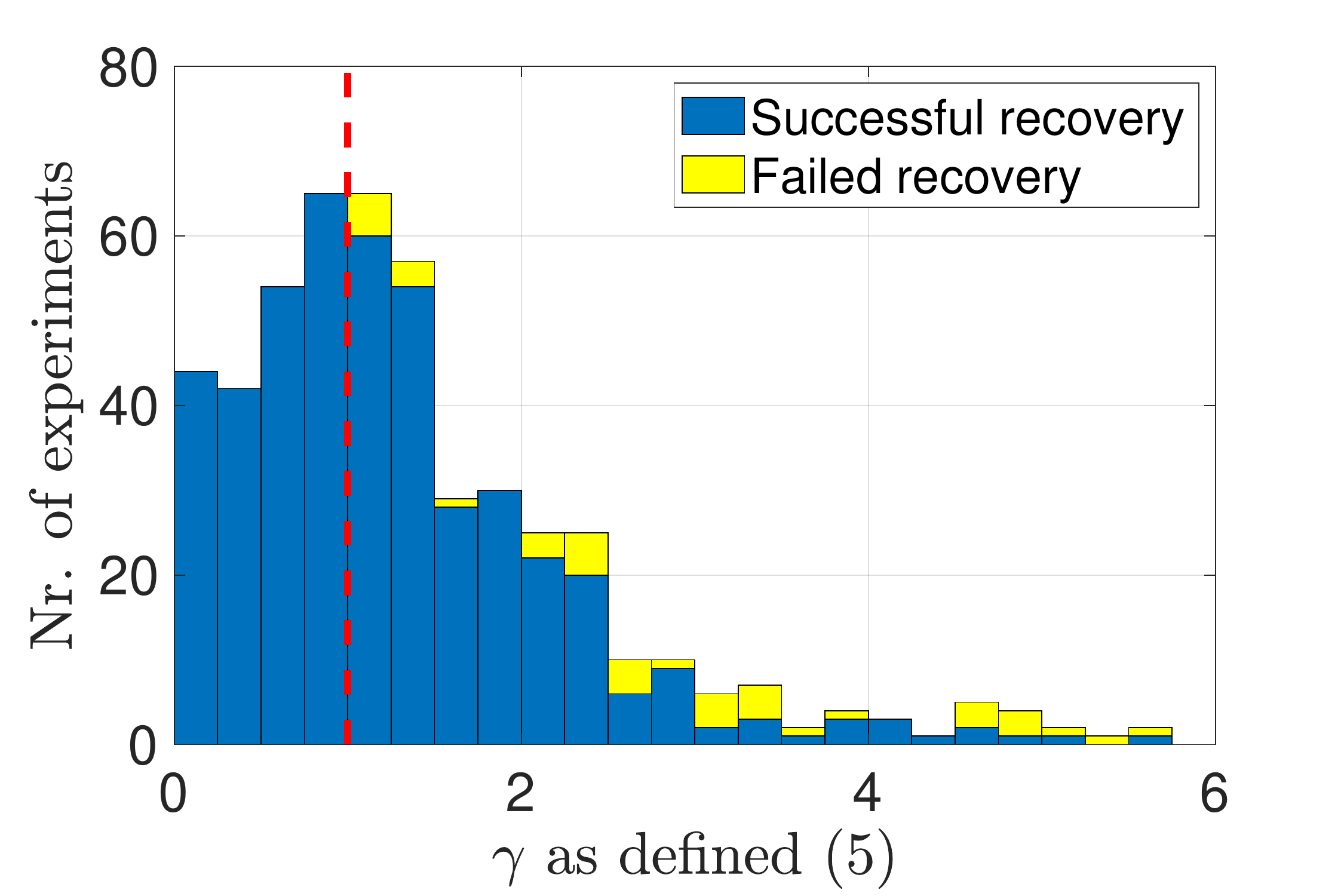}
		\centering{\small (a)}
	\end{minipage}%
	\begin{minipage}[c]{.25\textwidth}
		\includegraphics[width=\textwidth]{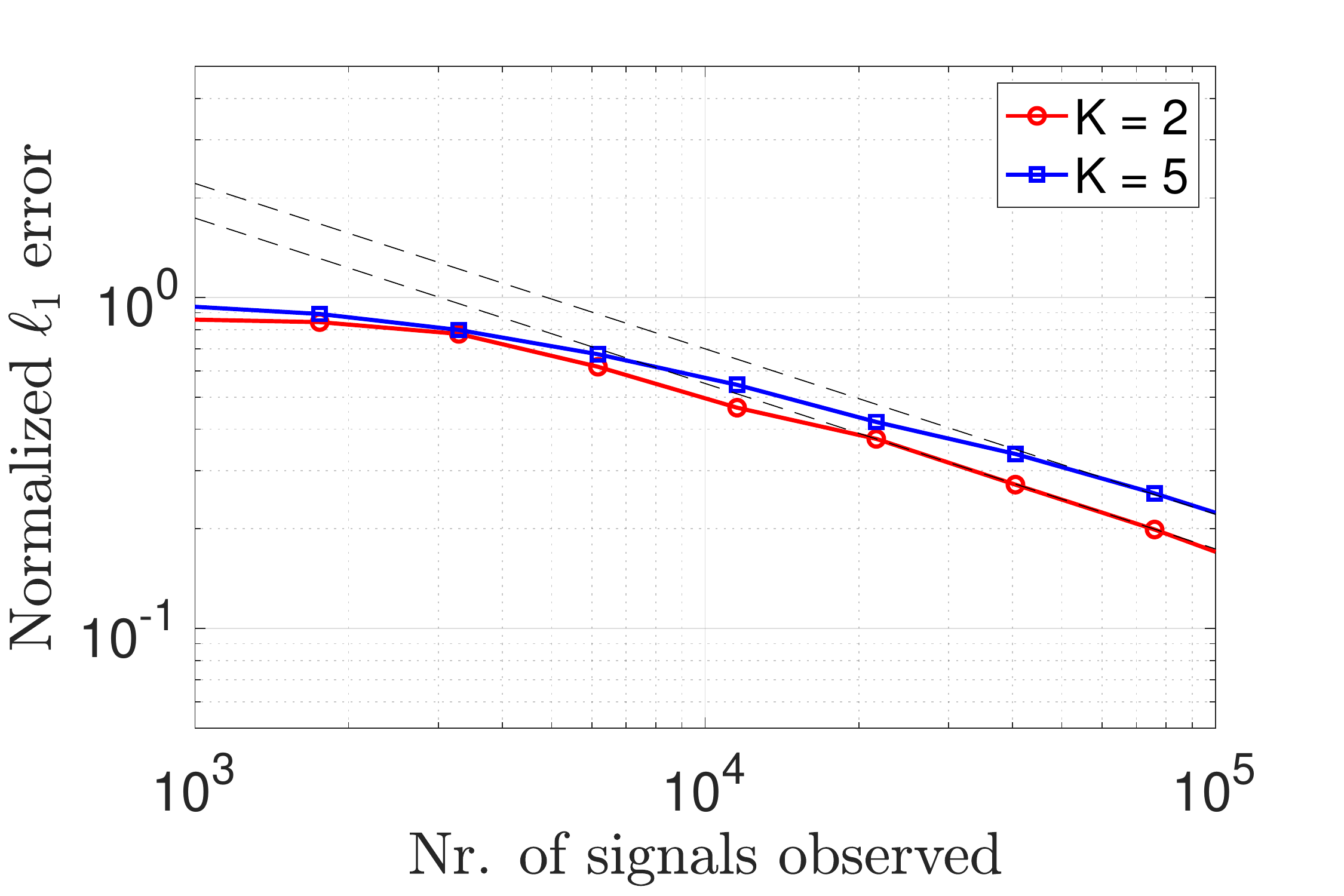}
		\centering{\small (b)}
	\end{minipage}%
	\begin{minipage}[c]{.25\textwidth}
		\includegraphics[width=\textwidth]{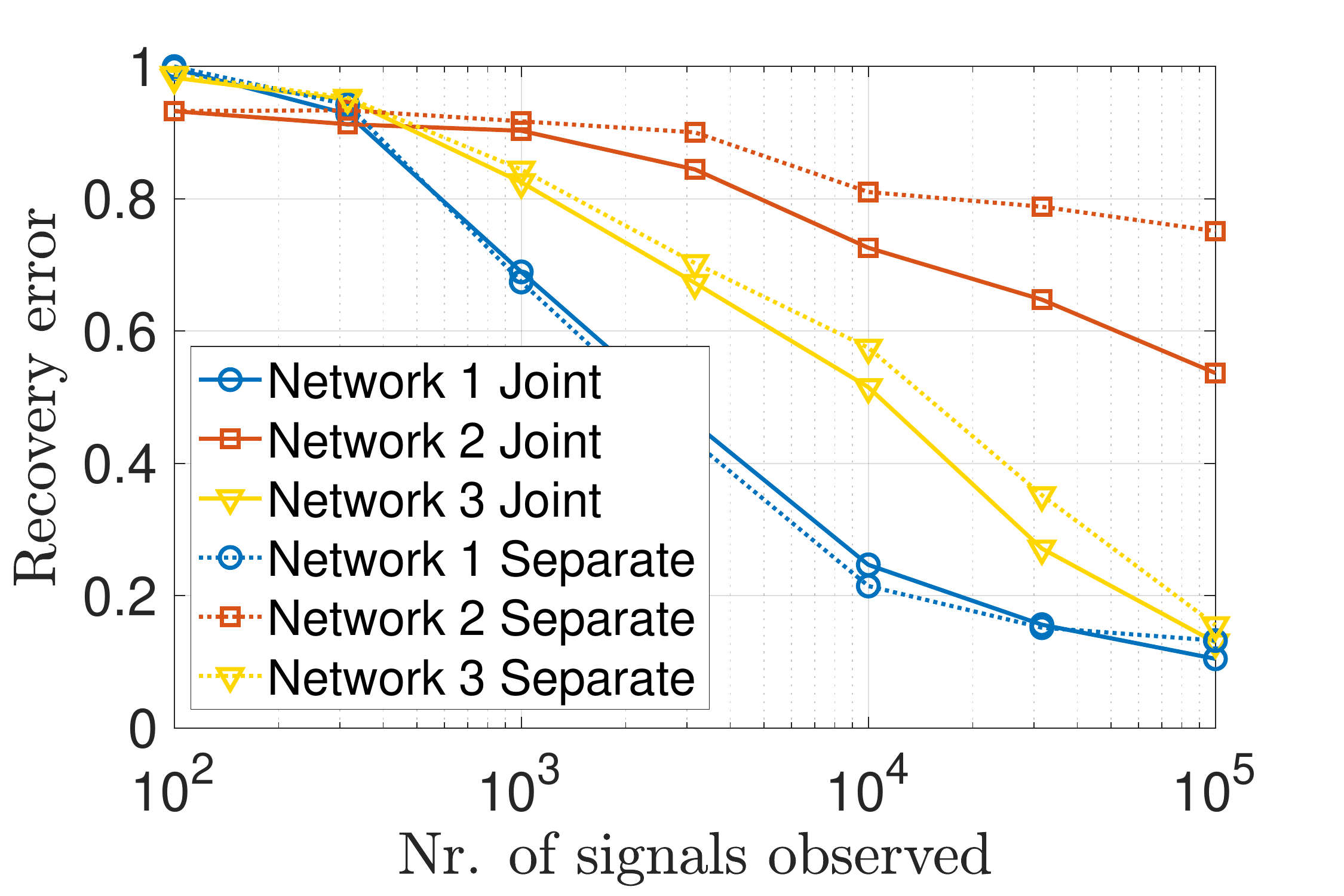}
		\centering{\small (c)}
	\end{minipage}%
	\begin{minipage}{.25\textwidth}
		\includegraphics[width=\textwidth]{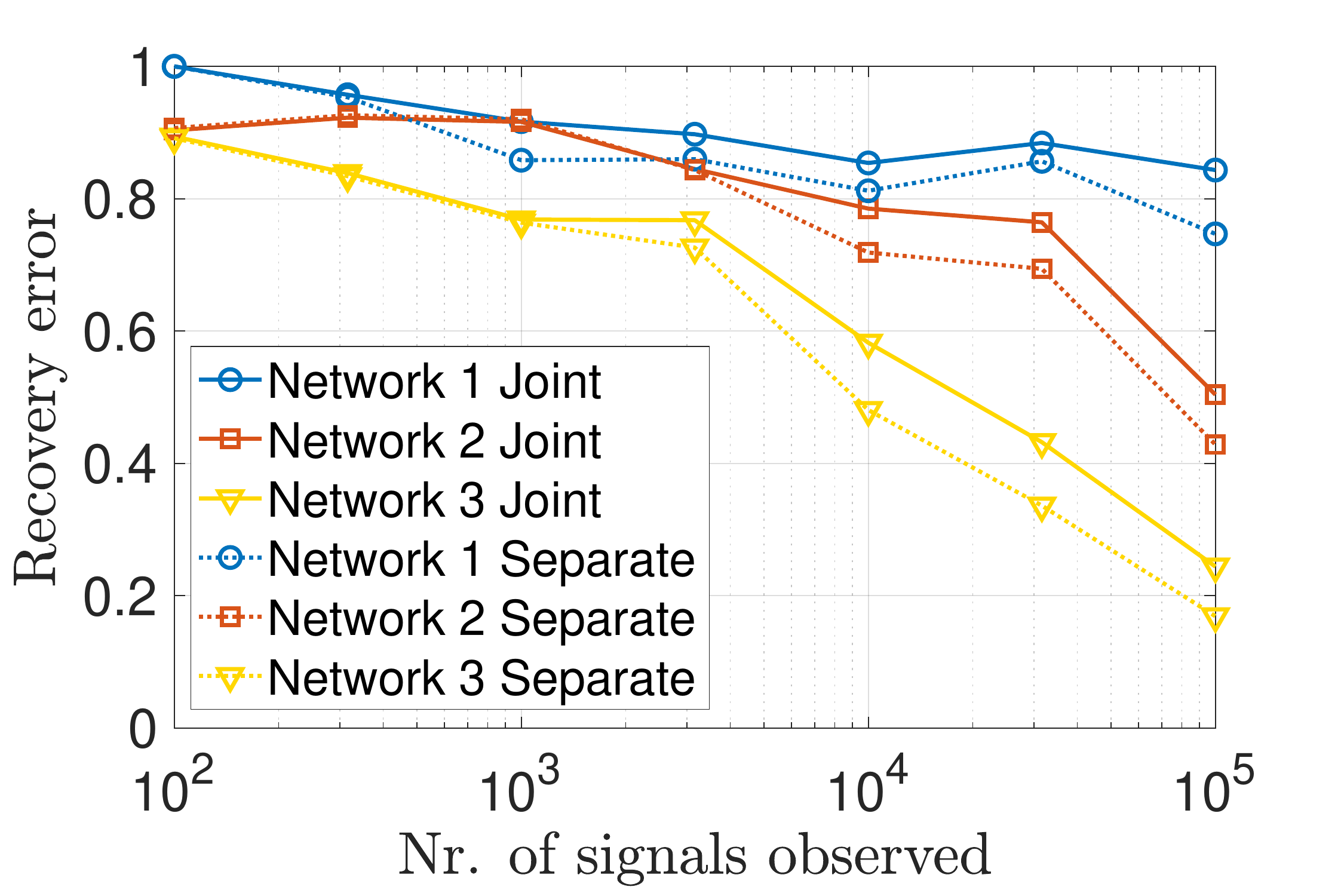}
		\centering{\small (d)}
	\end{minipage}
	
	\caption{\small (a)~Experimental validation of Theorem~\ref{T:theo_perfect_recovery}. For every realization where $\gamma$ in~\eqref{E:condition_recovery_noiseless} is strictly less than 1, perfect recovery is achieved.
	(b)~Experimental validation of Theorem~\ref{T:robust_recovery}. The sum of the $\ell_1$-norm recovery errors decreases as $1/\sqrt{n}$ as larger numbers of signals are observed. 
	(c)~Recovery error for three social networks with similar structure as a function of the number of signals in the computation of the sample covariance. The joint inference method in \eqref{E:l1_norm} achieves lower overall error than the separate inference of each network. (d)~Recovery error for three networks with no common structure as a function of the number of signals in the computation of the sample covariance. By enforcing a non-existent similarity between networks, the joint inference method underperforms compared to the separate inference.
	}
	
	\label{F:num_exp}
\end{figure*}
%%%%%%%%%%%%%%%%%%%%%%

Through synthetic and real-world graphs, we validate our theoretical claims and illustrate the performance of the proposed method for joint inference of networks.
	
\vspace{0.05in}
\noindent\textbf{Conditions for noiseless recovery.} 
In this numerical experiment, we illustrate the theoretical guarantees in Theorem~\ref{T:theo_perfect_recovery} by jointly inferring pairs of networks from perfect knowledge of the covariances of graph stationary processes.
More specifically, we generate 500 pairs of graphs where one graph in each pair is generated from an Erd\H{o}s-R\'enyi model \cite{bollobas1998random} of size $N \!=\! 20$ and edge-formation probability $p = 0.1$, and the other graph is obtained by randomly rewiring $3$ edges of the first one. 
Notice that this procedure ensures that both graphs in each pair are similar, thus motivating our joint inference method.
Covariance matrices of stationary processes in each graph are generated randomly by constructing filters $\bbH$ (see  Section~\ref{Ss:fund_GSP}) of size $L=3$ with normally distributed coefficients, and then setting $\bbC_\bbx = \bbH \bbH^\top$.
Our goal is to recover the adjacency matrices of each pair of graphs by solving 500 instances of~\eqref{eqn_one_norm}, where we set $\alpha_1 = \alpha_2 = \beta_{1,2}=1$. 
For each of these 500 attempts we record whether the recovery was successful or not, whether condition 1 in Theorem~\ref{T:theo_perfect_recovery} was satisfied or not -- it was satisfied in all cases --, and the value of $\gamma$ in~\eqref{E:condition_recovery_noiseless}. 
In Fig.~\ref{F:num_exp}a we plot the histogram of $\gamma$ discriminating by recovery performance. The figure clearly depicts the result of Theorem~\ref{T:theo_perfect_recovery} in that, for all cases in which $\gamma < 1$, relaxation \eqref{eqn_one_norm} achieves perfect recovery.
Equally important, Fig.~\ref{F:num_exp}a reveals that the bound stated in \eqref{E:condition_recovery_noiseless} is tight since some realizations with $\gamma = 1$ led to failed recoveries as indicated by the yellow portion of the bar to the right of the dashed line.

\vspace{0.05in}
\noindent\textbf{High-probability error bound.} 
We next demonstrate the upper bound for the recovery error in Theorem~\ref{T:robust_recovery} in the case of fixed nodes $N$ and graphs $K$. 
We generate one graph from an Erd\H{o}s-R\'enyi model with $N=20$ and edge-formation probability $p=0.4$, and the remaining $K-1$ graphs are obtained by rewiring the edges of the first graph with probability $q=0.3$. 
Graph filters are constructed as in the previous experiment. 
In this case, we are demonstrating robust graph inference using sample covariance matrices obtained from an increasing number of observed graph signals. 
Since $K$ and $N$ are fixed, we would expect the sum of the $\ell_1$-norm recovery errors to be upper bounded by $C/\sqrt{n}$ for a proper choice of the constant $C$ [cf.~\eqref{E:theorem_robust}], where we are neglecting the effect of $n$ on $\sigma_{\min}(\bbM)$.
Importantly, the square-root decrease in error should be more conspicuously observed for large $n$, since in this regime the dependence of $\sigma_{\min}(\bbM)$ on the number of observations is indeed negligible.
This is observed in practice, as portrayed in Fig.~\ref{F:num_exp}b, where we plot the recovery error of $K \in \{2,5\}$ graphs as the number of signals increases.
The error shown in the figure corresponds to the sum in \eqref{E:theorem_robust} normalized by the $\ell_1$-norm sum of the true GSOs, i.e., $\sum_{k=1}^K \|\hat{\bbS}^{(k)} - {\bbS^{(k)}}^* \|_1 / \sum_{k=1}^K \|{\bbS^{(k)}}^*\|_1$.
In both cases, the normalized $\ell_1$-norm error remains below the $C/\sqrt{n}$ bound, where $C$ is chosen for best fit, and the bound becomes especially tight for large values of $n$.

\vspace{0.05in}
\noindent\textbf{Joint inference of social networks.} Consider three graphs defined on a common set of nodes representing 32 students from the University of Ljubljana in Slovenia. 
The networks encode different types of interactions among the students, and were built by asking each student to select a group of preferred college mates for a number of situations, e.g., to discuss a personal issue or to invite to a birthday party\footnote{Access to the data and additional details are available at \url{http://vladowiki.fmf.uni-lj.si/doku.php?id=pajek:data:pajek:students}}.
The considered graphs are unweighted and symmetric, and the edge between $i$ and $j$ exists if either student $i$ picked $j$ in the questionnaire or vice versa.
Notice that the obtained networks are naturally similar to each other since the choices of friends across different situations do not vary greatly.
We test the recovery performance of the robust formulation in \eqref{E:l1_norm} where the sample covariances are estimated from varying numbers of graph signals, and $\epsilon_n$ is chosen as small as possible while ensuring feasibility.
The graph signals are synthetically generated following covariance matrices obtained from the GSOs as explained in the previous numerical experiment.
Fig.~\ref{F:num_exp}c portrays the joint recovery errors for the three networks as the number of observed signals varies, and compares them to the corresponding errors obtained from inferring the networks separately.
The error of an estimator $\hbS$ is quantified as $\| \bbS - \hbS\|_F / \| \bbS \|_F$, where $\bbS$ denotes the true GSO.
First notice that for an increasing number of observed signals we see a monotonous decrease in recovery error.
For instance, when going from $10^3$ to $10^4$ observations the error when inferring Network 1 is (approximately) divided by three.
This is expected since a larger number of observations entails a more reliable estimate of the covariance matrix.
More interestingly, we see an overall positive effect of the \emph{joint} inference compared to the corresponding separate inferences.
This effect is more conspicuous for Network 2, for which the inference based exclusively on its sample covariance has proven to be more challenging.

Finally, we repeat the above experiment but for three synthetically generated networks that model a scenario where the students choose their college mates completely at random. 
In this way, the similarity across the networks to be inferred is lost. 
Consequently, imposing this similarity in the joint inference problem is actually detrimental to the recovery performance as depicted in Fig.~\ref{F:num_exp}d. 
Indeed, from the figure it can be seen that for the three networks and for almost any possible number of observed signals the separate inference outperforms the joint method.

%%%%%%%%%%%%%%%%%%%%%%
% FIGURE
\begin{figure*}[!t]
	\centering
	
	\begin{minipage}[c]{.31\textwidth}
		\includegraphics[width=\textwidth]{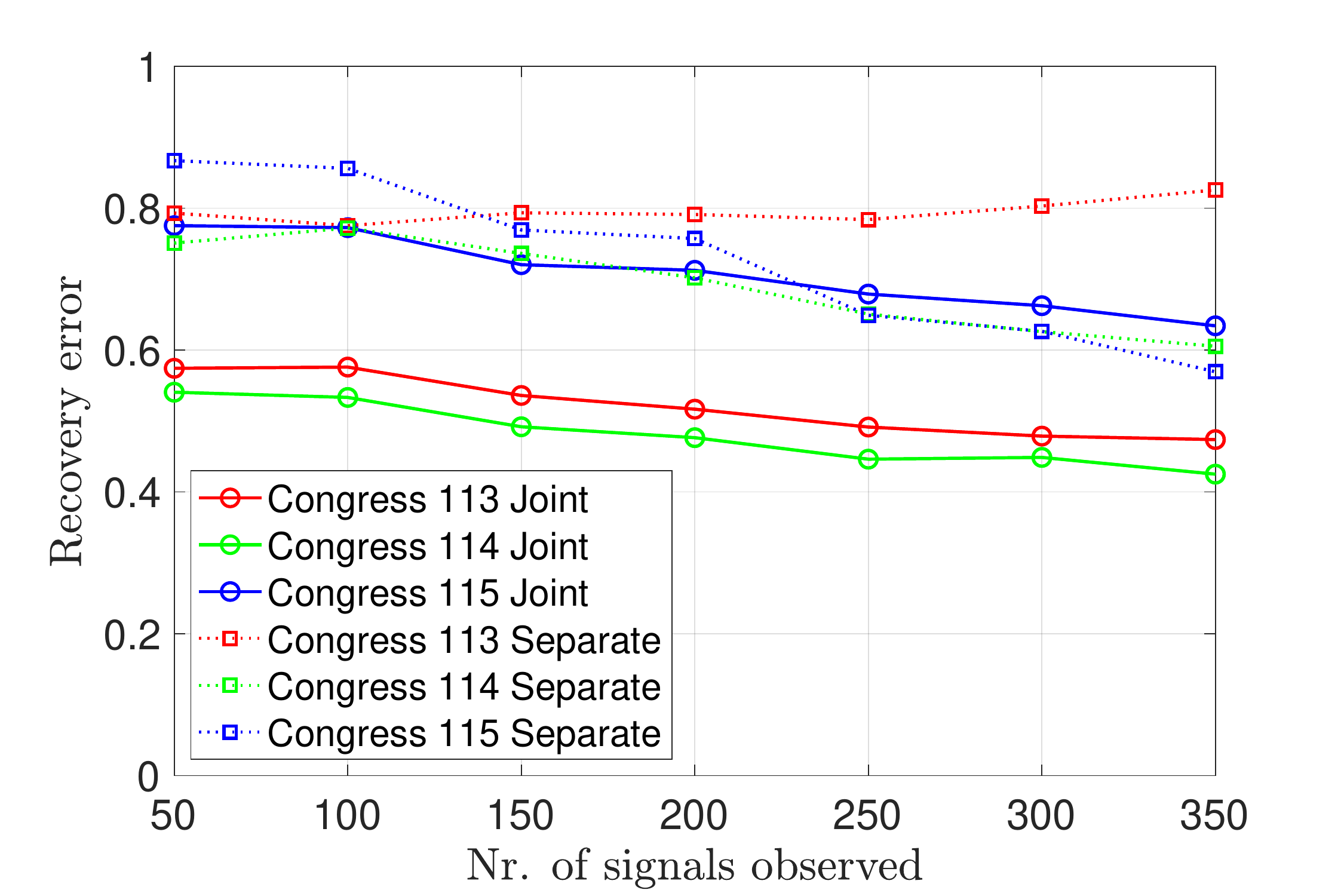}
		\centering{\small (a)}
	\end{minipage}%
	\begin{minipage}[c]{.23\textwidth}
		\includegraphics[width=\textwidth]{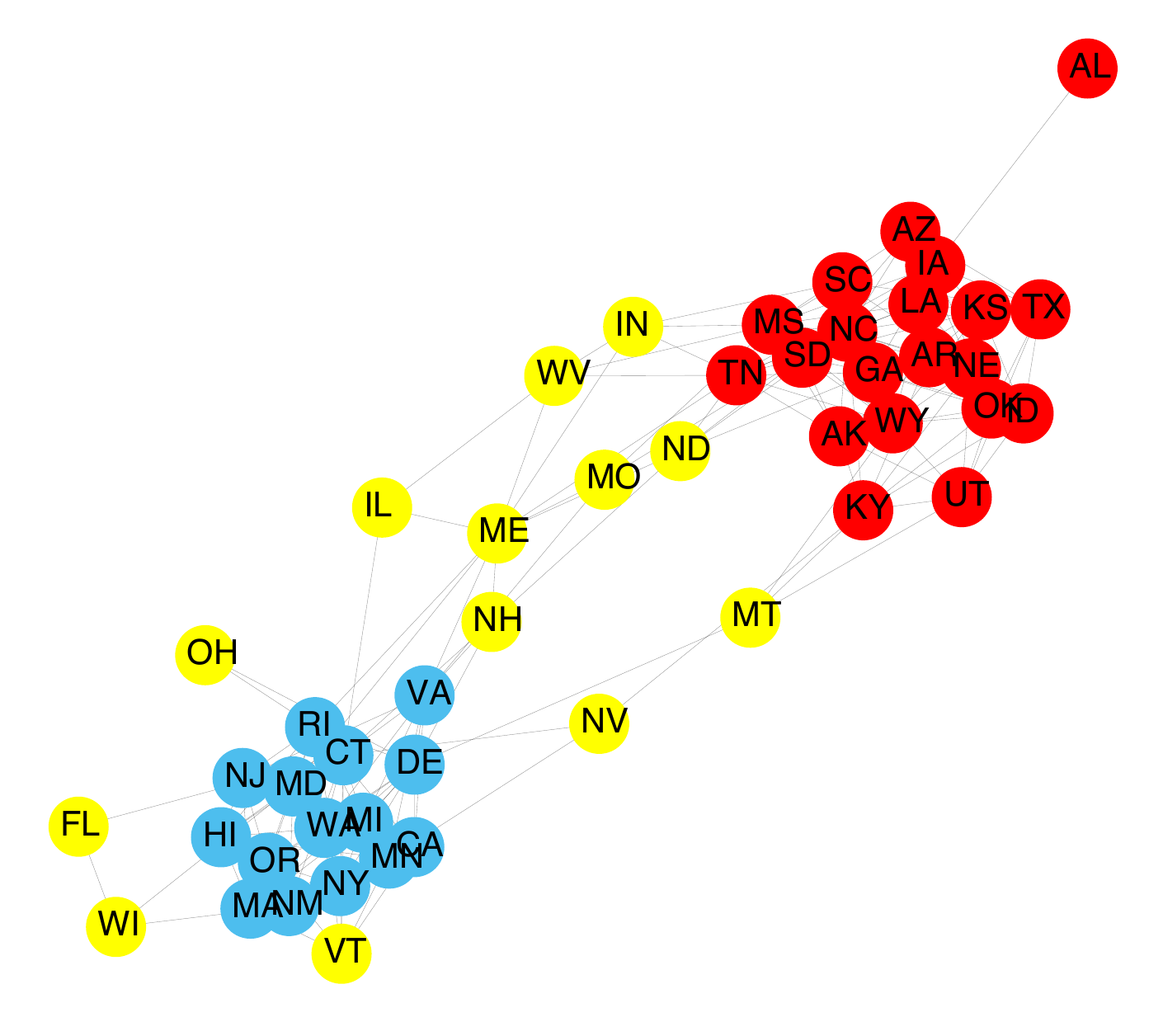}
		\centering{\small (b) True graph}
	\end{minipage}%
	\begin{minipage}[c]{.23\textwidth}
		\includegraphics[width=\textwidth]{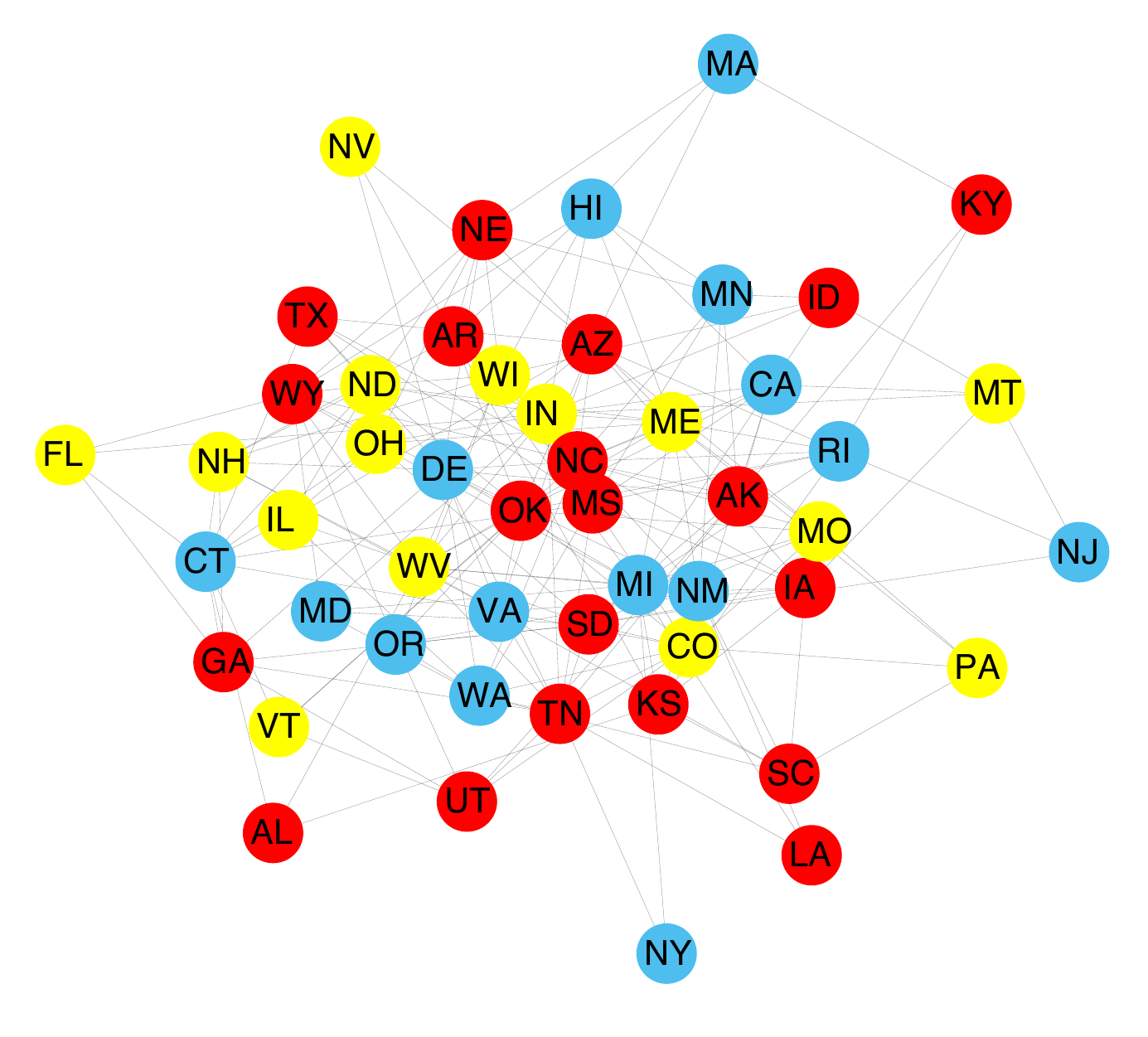}
		\centering{\small (c) Separately inferred}
	\end{minipage}%
	\begin{minipage}{.23\textwidth}
		\includegraphics[width=\textwidth]{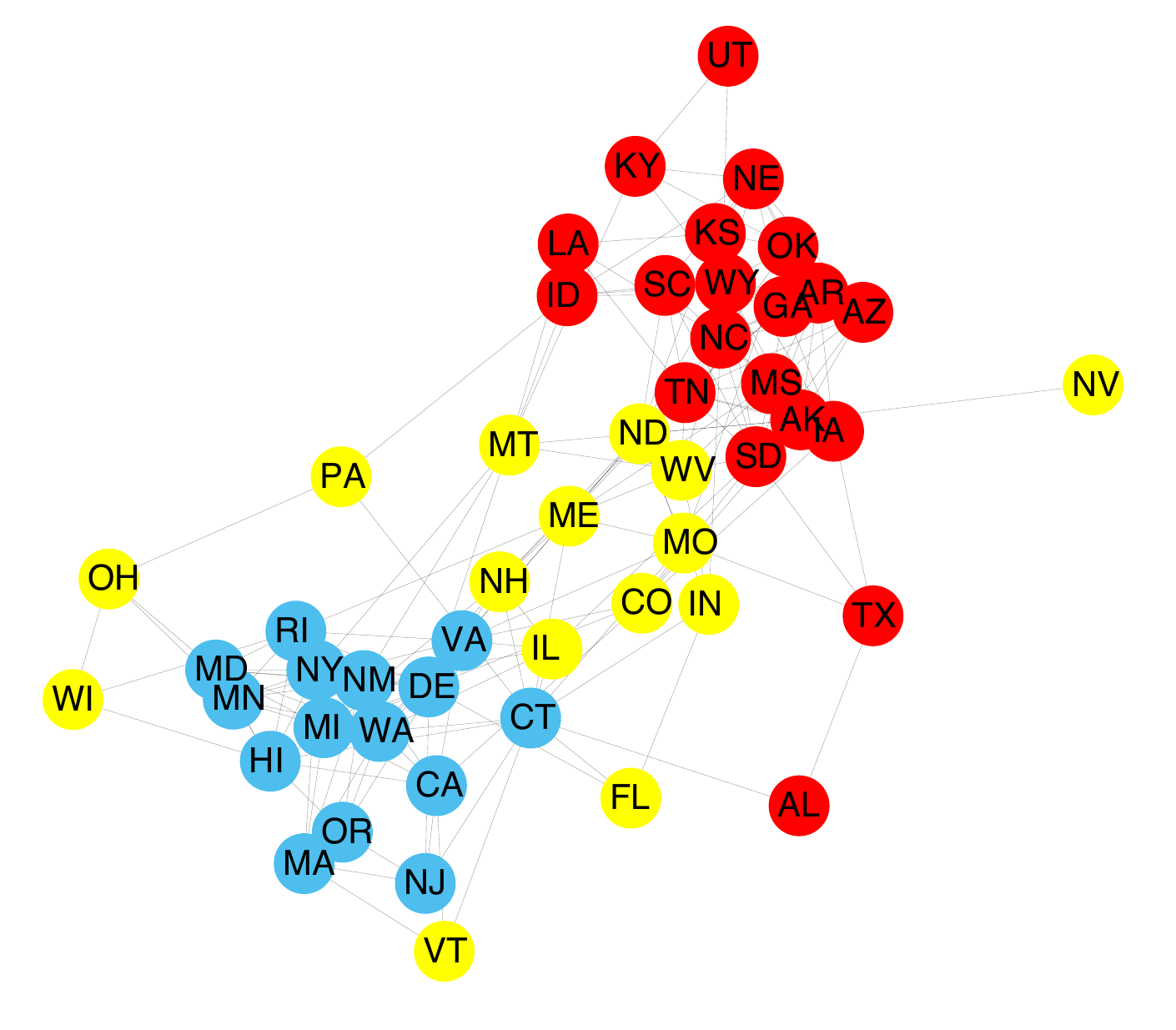}
		\centering{\small (d) Jointly inferred}
	\end{minipage}
	
	\caption{\small (a)~ Mean recovery error over ten trials for three senate networks as a function of the number of signals considered in the computation of the sample covariance. Joint inference demonstrates less overall recovery error than separate inference of each senate network.
	(b)~ True graph of senate network for 114th congress with top-200 edges sorted by weight. Red nodes correspond to states with two Republican senators, blue nodes with two Democratic senators, and yellow nodes with senators from differing parties. 
		(c)~ Separately recovered senate network for 114th congress with top-200 edges sorted by weight. Network recovery for a limited number of signals shows a mixed structure when each network is estimated alone.
(d)~Jointly recovered senate network for 114th congress with top-200 edges sorted by weight. A more similar structure to the true graph is observed when joint inference is applied for three similar senate networks.
	}
	
	\label{F:sen_exp}
\end{figure*}
%%%%%%%%%%%%%%%%%%%%%%

\vspace{0.05in}
\noindent\textbf{Senate networks.} 
The comparison of joint and separate graph inference is also performed with real-world data of U.S. congress roll-call votes \cite{lewis2020voteview}. 
We observe the votes of 3 congresses, 113th (2013 to 2015), 114th (2015 to 2017), and 115th (2017 to 2019), from 2 senators per state (100 total). All $K=3$ congresses are represented as networks, where senator opinions per state are combined for $N=50$ nodes shared by each graph. Nodal values for each state consist of the sum of the votes of both senators, where yea, nay, and other cases (such as abstention) are represented by 1, -1, and 0, respectively. 
The total number of roll-calls (graph signals) for the 113th, 114th, and 115th congresses are respectively $657$, $502$, and $599$.
Each state is separated into one of three categories based on the party affiliation of its senators. States are labeled as (i) D if both senators are in the Democratic Party, (ii) R if both senators are in the Republican Party, and (iii) M if senators are from different (mixed) parties. 

In the absence of ground-truth senate networks, we deem as true underlying graphs those separately inferred for each congress when considering all the available graph signals.
Moreover, to recover graphs on which the observed signals are not only stationary but also smooth, we add a regularization term $\| \bbS \circ \bbZ \|$ to the optimization objective, where $\bbZ$ is the pairwise distance matrix $Z_{ij} = \| \bbx_i - \bbx_j \|_2^2$, and $\bbx_i$ contains the value of all signals at the $i$-th node; see~\cite{Kalofolias2016inference_smoothAISTATS16}.
Having established the ground-truth baselines, we perform joint and separate inference from limited observations and compare the estimation accuracy when gradually increasing the number of signals considered in the covariance computation. 
Subsets are randomly selected from all available votes, and ten trials of randomized subsets are performed to observe their mean behavior; see Fig.~\ref{F:sen_exp}a. 

Although the true networks were inferred separately, joint inference of senate networks markedly outperforms separate inference when a limited number of signals are available. 
This further reinforces our intuition that pooling observations from similar networks is especially relevant in data scarce settings.
To better illustrate the difference in the inferred networks, in Figs.~\ref{F:sen_exp}b through \ref{F:sen_exp}d we provide spring layout plots of the true, separately inferred, and jointly inferred networks for congress 114th when 350 signals are observed. 
For clarity, only the top-200 edges sorted by weight are drawn.
From the figures it becomes evident that the joint inference helps preserve the partisan structure of the true network whereas this important network feature is not recovered when performing separate inference.

%%%%%%%%%%%%%%%%%%%%%%%%%%%%%%%%%%%%%%%%%%%%%%%%%%%%%%%%%%%%%%
% CONCLUSIONS
\section{Conclusions}\label{S:conclusions}
We presented a method for jointly estimating multiple graphs from observed graph signals. 
The inference task was posited as a sparse recovery optimization problem regularized by the differences between the recovered graphs and subject to algebraic constraints derived from the assumption that the observed signals are stationary on the underlying graphs.
A convex relaxation of the aforementioned optimization problem was presented and its tightness was shown under sufficient conditions for the case of perfect knowledge on the signal covariances.
Furthermore, for the more relevant case where the covariances are estimated, a robust variation of the optimization problem is presented along with a high-probability bound on the recovery error.
Finally, the results and intuition discussed throughout the paper were illustrated via numerical experiments on synthetic and real-world data.

Regarding potential avenues for future research, two generalizations of the setting here presented are of special interest.
1) It would be of interest to relax the assumption that we know on which graph each signal is defined. 
This case would require clustering the signals based on their estimated source graph and, most probably, an iterative formulation where the graphs are inferred and the signals reassigned between the graphs until convergence.
2) It would be of interest to consider setups where the node sets (and their cardinality) are not the same across the graphs to be inferred.
The rising popularity of graphons~\cite{avella_2020_centrality} may contribute to solving this setting, where the association of multiple graphs with a graphon presents a potential direction for joint graph inference with the underlying similarity between graphs dictated by the probability of being generated by a common graphon.

%%%%%%%%%%%%%%%%%%%%%%%%%%%%%%%%%%%%%%%%%%%%%%%%%%%%%%%%%%%%%%
% APPENDICES
\appendices

%%%%%%%%%%%%%%%%%%%%%%%%%%%%%%%%%%%%%%%%%%%%%%%%%%%%%%%%%%%%%%
% PROOF OF CLAIMS
\section{Proofs of Claims~\ref{C:problem_q} and~\ref{C:reqs_3}}

We first state the following two lemmas that will be used to prove Claim~\ref{C:reqs_3}.

%%%%%%%%%%%
% LEMMA 1
\begin{mylemma}\cite[Lemma 2]{cai2016joint}\label{L:prob_bound_norm}
	Suppose $\bbr_1, \cdots, \bbr_n$ are $K$-dimensional random vectors satisfying $\E{\bbr_i} = \mathbf{0}$ and $\Vert \bbr_i \Vert_2 \leq M$ for $1 \leq i \leq n$. We have for any $s > 0$ and $r > s$
	\begin{equation}
	\begin{aligned}
	\!\!&\! \mathbb{P} \left(\left\| \sum_{i=1}^n \bbr_i \right\|_2 \geq r\right) \leq \mathbb{P} \left( \| \bbz \|_2 \geq (r - s) / \lambda_{\max}^{1/2} \right) + L,
%	\!\!&\! \mathbb{P} (\| \sum_{i=1}^n \bbr_i \|_2 \geq r) \leq \mathbb{P} \left( \| \bbz \|_2 \geq (r - s) / \lambda_{\max}^{1/2} \right) + L, \\
%	\!\!&\! \text{where} \,\,\, L = c_1 K^{5/2} \exp\left(-\frac{c_2  s}{K^{5/2} M}\right)
%	\mathbb{P} (\| \sum_{i=1}^n \bbr_i \|_2 \geq r) \leq \mathbb{P} \left( \| \bbz \|_2 \geq (r - s) / \lambda_{\max}^{1/2} \right) + c_1 K^{5/2} \exp(-c_2 K^{-5/2} s / M),
	\end{aligned}
	\end{equation}
	where $L = c_1 K^{5/2} \exp(-c_2 K^{-5/2} s / M)$, $\lambda_{\max}$ is the largest eigenvalue of $\textrm{Cov}(\sum_{i=1}^n \bbr_i)$, $\bbz$ is a $K$-dimensional standard normal random vector and $c_1, c_2$ are positive constants.
\end{mylemma}
%%%%%%%%%%%

%%%%%%%%%%%
% LEMMA 2
\begin{mylemma}\label{L:tail_bounds}
	Denoting by $a^{(k)}$ independent realizations of the random variable $a \sim \mathcal{N}(0, \sigma^2)$, the following tail bound holds
	\begin{equation}~\label{lem:chisq}
	\mathbb{P}\left( \frac{1}{m} \sum_{k=1}^m (a^{(k)})^2 - \mathbb{E} [ a^2 ] \geq \sigma^2 \, t \right) \leq \exp\left( -\frac{m}{8} \min(t^2, t) \right).
	\end{equation}
\end{mylemma}
%%%%%%%%%%%

Lemma~\ref{L:prob_bound_norm} bounds in probability the sum of the norm of bounded random vectors whereas Lemma~\ref{L:tail_bounds} is a standard result about tail bounds of chi-squared random variables. Having introduced these results, we can now show the two claims.

%%%%%% PROOF OF CLAIM 1 %%%%%%%%%%%%%%%%%%%%%%%%%%%%%
\subsection{Proof of Claim~\ref{C:problem_q}} 

This proof has been partially inspired by \cite[Theorem 2]{zhang2016one}. We are first going to show that condition \emph{l1)} guarantees the existence of a vector $\bby \in \reals^{\left(K + \binom{K}{2} \right) N^2}$ -- that will be denominated \emph{dual certificate} -- such that $\bbR^\top \bby \in \mathrm{Im}(\bbPhi_q^\top)$, $\bby_\ccalK = \mathrm{sign}( \bbR_\ccalK \bbs_\ccalL^* )$, and $\| \bby_{\ccalK^c} \|_\infty < 1$. In fact, we show here that we may attain that $\bby_{\ccalK^c} = \mathbf{0}$. Indeed, consider the vector $\bby$ given by
\begin{equation}\label{E:main_lemma_010}
\bby =  \bbI_{\ccalK}^\top \, \mathrm{sign}( \bbR_\ccalK \bbs_\ccalL^* ).
\end{equation}

That $\bby_\ccalK = \mathrm{sign}( \bbR_\ccalK \bbs_\ccalL^* )$, and $\| \bby_{\ccalK^c} \|_\infty = 0$ follow immediately from \eqref{E:main_lemma_010}. Moreover, we have that $ \bbR^\top \bby \in \mathrm{Im}(\bbPhi_q^\top)$ by realizing that $\bbR^\top \bby = \bbPhi_q^\top  \bbPhi_q (\bbPhi_q^\top  \bbPhi_q)^{-1} \bbR^\top \, \bbI_{\ccalK}^\top \, \mathrm{sign}( \bbR_\ccalK \bbs_\ccalL^* )$, where condition \emph{l1)} guarantees the existence of the inverse. Now that we have established the existence of the dual certificate $\bby$, it is helpful to notice that $\| \bbR \bbs^*_{\ccalL} \|_1 = \bby^\top \bbR \bbs^*_{\ccalL}$. 

We are ready to show the bound in \eqref{E:def_c_q}. Consider an arbitrary vector $\bbu \in \reals^{\left(K + \binom{K}{2} \right) N^2}$ such that $\mathrm{supp}(\bbu) \subseteq \ccalK$. 
Letting $\bbrho = \bbR \hat{\bbs}_{\ccalL}^{(q)} - \bbR \bbs^*_{\ccalL}$, $\bbrho_1 = \bbR\hat{\bbs}_{\ccalL}^{(q)} - \bbu$, and $\bbrho_2 = \bbR \bbs^*_{\ccalL} - \bbu$, we have that

\begin{equation}\label{E:main_lemma_030}
\| \bbrho \|_1 \leq \| \bbrho_1 \|_1 + \| \bbrho_2 \|_1.
%\| \bbR \hat{\bbs}_{\ccalL}^{(q)} - \bbR \bbs^*_{\ccalL} \|_1 \leq \| \bbR\hat{\bbs}_{\ccalL}^{(q)} - \bbu \|_1 + \| \bbR \bbs^*_{\ccalL} - \bbu \|_1.
\end{equation}

We first focus on bounding the second summand in \eqref{E:main_lemma_030}. By leveraging the fact that the support of $ \bbrho_2$ is contained in $\ccalK$ we may write that
%We first focus on bounding the second summand in \eqref{E:main_lemma_030}. By leveraging the fact that the support of $ \bbR \bbs^*_{\ccalL} - \bbu$ is contained in $\ccalK$ we may write that

%
\begin{equation}
\begin{aligned}\label{E:main_lemma_040}
\| \bbrho_2 \|_1 & \leq &&\sqrt{|\ccalK|} \,\, \| \bbrho_2 \|_2 \\
& \leq &&\sqrt{|\ccalK|} \,\, \| \bbrho  \|_2 + \sqrt{|\ccalK|} \,\, \| \bbrho_1 \|_2 \\
& \leq &&\sqrt{|\ccalK|} \,\, \sigma_{\max}(\bbR) \| \bbs^*_{\ccalL} - \hat{\bbs}_{\ccalL}^{(q)}  \|_2 + \sqrt{|\ccalK|} \,\, \| \bbrho_1 \|_1 \\
& \leq &&\frac{\sqrt{|\ccalK|} \sigma_{\max}(\bbR)}{\sigma_{\min}(\bbPhi_q)} \,\, \| \bbPhi_q (\bbs^*_{\ccalL} - \hat{\bbs}_{\ccalL}^{(q)} ) \|_2 + \sqrt{|\ccalK|} \,\, \| \bbrho_1 \|_1, 
%\!\!&\! \| \bbR \bbs^*_{\ccalL} - \bbu \|_1 \leq \sqrt{|\ccalK|} \,\, \| \bbR \bbs^*_{\ccalL} - \bbu \|_2 \\
%& \quad \leq \sqrt{|\ccalK|} \,\, \| \bbR( \bbs^*_{\ccalL} - \hat{\bbs}_{\ccalL}^{(q)})  \|_2 + \sqrt{|\ccalK|} \,\, \| \bbR \hat{\bbs}_{\ccalL}^{(q)}  - \bbu \|_2 \\
%& \quad \leq \sqrt{|\ccalK|} \,\, \sigma_{\max}(\bbR) \| \bbs^*_{\ccalL} - \hat{\bbs}_{\ccalL}^{(q)}  \|_2 + \sqrt{|\ccalK|} \,\, \| \bbR \hat{\bbs}_{\ccalL}^{(q)}  - \bbu \|_1 \\
%& \quad \leq  \frac{\sqrt{|\ccalK|} \sigma_{\max}(\bbR)}{\sigma_{\min}(\bbPhi_q)} \,\, \| \bbPhi_q (\bbs^*_{\ccalL} - \hat{\bbs}_{\ccalL}^{(q)} ) \|_2 + \sqrt{|\ccalK|} \,\, \| \bbR \hat{\bbs}_{\ccalL}^{(q)}  - \bbu \|_1, 
\end{aligned}
\end{equation}
where in the third inequality we used that for an arbitrary vector $\bbx$ it holds that $\| \bbx \|_2 \leq \| \bbx \|_1$, and in the last inequality we used that $\| \bbx \|_2 \leq \| \bbA \bbx \|_2 / \sigma_{\min}(\bbA)$, for every full column rank matrix $\bbA$. Condition \emph{l1)} guarantees the validity of this operation. 

We now find an upper bound for $\| \bbrho_1 \|_1$ for the vector $\bbu$ that minimizes this norm. 
More precisely, we want to bound $\xi:= \min_{\bbu | \mathrm{supp}(\bbu) \subseteq \ccalK} \| \bbrho_1 \|_1$. We may rewrite the support constraint on $\bbu$ as $\bbI_{\ccalK^c} \bbu = \mathbf{0}$. Thus, the Lagrangian $L(\bbu, \bbv)$ of the minimization problem becomes
\begin{equation}
\begin{aligned}\label{E:main_lemma_060}
L(\bbu, \bbv) &= \| \bbrho_1 \|_1 + \bbv^\top \bbI_{\ccalK^c} \bbu \\
\quad &= \| \bbrho_1 \|_1 + \bbv^\top \bbI_{\ccalK^c}  ( \bbu - \bbR \hat{\bbs}_{\ccalL}^{(q)} ) + \bbv^\top \bbI_{\ccalK^c} \bbR \hat{\bbs}_{\ccalL}^{(q)}.
\end{aligned}
\end{equation}
From duality theory we have that $\xi = \max_\bbv \min_\bbu L(\bbu, \bbv)$. Moreover, if we define $\bbw := \bbI_{\ccalK^c}^\top \bbv$, we have that
\begin{align}\label{E:main_lemma_070}
\xi = \,\, \max_{\bbw | \mathrm{supp}(\bbw) \subseteq \ccalK^c}  \min_\bbu \,\, \| \bbrho_1 \|_1 + \bbw^\top ( \bbu - \bbR \hat{\bbs}_{\ccalL}^{(q)} ) + \bbw^\top \bbR \hat{\bbs}_{\ccalL}^{(q)}.
\end{align}
By minimizing with respect to $\bbu$, for \eqref{E:main_lemma_070} not to result in $- \infty$, it must be that $\| \bbw \|_\infty \leq 1$. 
Otherwise, if $|w_r|>1$ for some index $r$, the corresponding entry $u_r$ can take a $- \infty$ value resulting in an unbounded minimization of $\xi$. In the case where $\| \bbw \|_\infty \leq 1$, the minimum for $\bbu$ is attained when $\bbu = \bbR \hat{\bbs}_{\ccalL}^{(q)}$. It thus follows that
\begin{align}\label{E:main_lemma_080}
\xi = \,\, \max_{\bbw | \, \mathrm{supp}(\bbw) \subseteq \ccalK^c, \, \| \bbw \|_\infty \leq 1} \bbw^\top  \bbR \hat{\bbs}_{\ccalL}^{(q)}.
\end{align}
Recalling that $\bby$ is the previously introduced dual certificate [cf.~\eqref{E:main_lemma_010}], we may write that
\begin{align}\label{E:main_lemma_090}
\xi = \,\,  \max_{\bbw | \, \mathrm{supp}(\bbw) \in \ccalK^c, \, \| \bbw \|_\infty \leq 1} (\bby + \bbw)^\top  \bbR \hat{\bbs}_{\ccalL}^{(q)} - \bby^\top \bbR \hat{\bbs}_{\ccalL}^{(q)}. 
\end{align}
Moreover, since $\| \bby \|_\infty = 1$, $\| \bbw \|_\infty \leq 1$ and the supports of $\bby$ and $\bbw$ do not intersect, it readily follows that $\| \bby + \bbw \|_\infty \leq 1$. Consequently, $(\bby + \bbw)^\top  \bbR \hat{\bbs}_{\ccalL}^{(q)} \leq \| \bbR \hat{\bbs}_{\ccalL}^{(q)} \|_1$. By substituting this in \eqref{E:main_lemma_090} we obtain that
\begin{align}\label{E:main_lemma_100}
\xi \leq \| \bbR \hat{\bbs}_{\ccalL}^{(q)} \|_1 -  \bby^\top  \bbR \hat{\bbs}_{\ccalL}^{(q)} .
\end{align}
Leveraging the fact that $\| \bbR \bbs^*_{\ccalL} \|_1 = \bby^\top \bbR \bbs^*_{\ccalL}$, we may write
\begin{equation}
\begin{aligned}\label{E:main_lemma_110}
& \xi \!\!\! &&\leq \| \bbR \hat{\bbs}_{\ccalL}^{(q)} \|_1 - \| \bbR \bbs^*_{\ccalL} \|_1 + \bby^\top \bbR ( \bbs^*_{\ccalL} - \hat{\bbs}_{\ccalL}^{(q)} ) \\
& &&\leq \bby^\top \bbR( \bbs^*_{\ccalL} - \hat{\bbs}_{\ccalL}^{(q)} ),
%& \|  \bbR \hat{\bbs}_{\ccalL}^{(q)} \|_1 -  \bby^\top \bbR \hat{\bbs}_{\ccalL}^{(q)} &&= \| \bbR \hat{\bbs}_{\ccalL}^{(q)} \|_1 - \| \bbR \bbs^*_{\ccalL} \|_1 + \bby^\top \bbR ( \bbs^*_{\ccalL} - \hat{\bbs}_{\ccalL}^{(q)} ) \\
%& &&\leq \bby^\top \bbR( \bbs^*_{\ccalL} - \hat{\bbs}_{\ccalL}^{(q)} ),
\end{aligned}
\end{equation}
where the inequality follows from $\| \bbR \hat{\bbs}_{\ccalL}^{(q)} \|_1 \leq  \| \bbR \bbs^*_{\ccalL} \|_1$ since $\hat{\bbs}_{\ccalL}^{(q)}$ is a minimizer of \eqref{E:proof_recovery_030} whereas $ \bbs^*_{\ccalL}$ is a feasible solution of \eqref{E:proof_recovery_030} due to condition \emph{l2)}. Lastly, since we know that $\bbR^\top \bby$ can be written as $\bbR^\top \bby = \bbPhi_q^\top  \bbPhi_q (\bbPhi_q^\top  \bbPhi_q)^{-1} \bbR^\top \, \bbI_{\ccalK}^\top \, \mathrm{sign}( \bbR_\ccalK \bbs_\ccalL^* )$, we may rewrite \eqref{E:main_lemma_110} as (recalling the definition of $\xi$)
\begin{align}\label{E:main_lemma_120}
&\|\bbrho_1\|_1 \!\!\!\! &&\leq  \mathrm{sign}( \bbR_\ccalK \bbs_\ccalL^* )^\top \bbI_{\ccalK} \bbR (\bbPhi_q^\top  \bbPhi_q)^{-1}  \bbPhi_q^\top \bbPhi_q ( \bbs^*_{\ccalL} - \hat{\bbs}_{\ccalL}^{(q)} ) \nonumber \\
& &&\leq \frac{\sqrt{|\ccalK|} \sigma_{\max}(\bbR)}{\sigma_{\min}(\bbPhi_q)}  \| \bbPhi_q ( \bbs^*_{\ccalL} - \hat{\bbs}_{\ccalL}^{(q)} )\|_2,
%\| \bbR \hat{\bbs}_{\ccalL}^{(q)} \|_1 -  \bby^\top \bbR \hat{\bbs}_{\ccalL}^{(q)} & \leq  \mathrm{sign}( \bbR_\ccalK \bbs_\ccalL^* )^\top \bbI_{\ccalK} \bbR (\bbPhi_q^\top  \bbPhi_q)^{-1}  \bbPhi_q^\top \bbPhi_q ( \bbs^*_{\ccalL} - \hat{\bbs}_{\ccalL}^{(q)} ) \nonumber \\
%& \leq \| \mathrm{sign}( \bbR_\ccalK \bbs_\ccalL^* )^\top \bbI_{\ccalK}\|_2 \|\bbR\|_2 \| (\bbPhi_q^\top  \bbPhi_q)^{-1}  \bbPhi_q^\top\|_2 \| \bbPhi_q ( \bbs^*_{\ccalL} - \hat{\bbs}_{\ccalL}^{(q)} )\|_2 \leq \frac{\sqrt{|\ccalK|} \sigma_{\max}(\bbR)}{\sigma_{\min}(\bbPhi_q)}  \| \bbPhi_q ( \bbs^*_{\ccalL} - \hat{\bbs}_{\ccalL}^{(q)} )\|_2,
\end{align}
where the second inequality follows from the fact that every positive scalar is equal to its $\ell_2$ norm. By substituting \eqref{E:main_lemma_120} back in \eqref{E:main_lemma_040} and then back in \eqref{E:main_lemma_030}, we obtain that
\begin{align}\label{E:main_lemma_130}
\| \bbrho \|_1 \leq \left( 2 + \sqrt{|\ccalK|} \right) \, \frac{\sqrt{|\ccalK|}\sigma_{\max}(\bbR)}{\sigma_{\min}(\bbPhi_q)}  \| \bbPhi_q ( \bbs^*_{\ccalL} - \hat{\bbs}_{\ccalL}^{(q)} )\|_2.
\end{align}
Two observations are sufficient to obtain \eqref{E:def_c_q} from \eqref{E:main_lemma_130}. First, notice that since $\bbs^*_{\ccalL}$ and $ \hat{\bbs}_{\ccalL}^{(q)}$ both belong to the feasibility set of \eqref{E:proof_recovery_030}, we must have that $\| \bbPhi_q (\bbs^*_{\ccalL} - \hat{\bbs}_{\ccalL}^{(q)} ) \|_2 \leq 2 \epsilon_n$. Second, from compatibility of matrix induced norms and the fact that $\bbR$ is full column rank we have that $\| \bbs^*_{\ccalL} - \hat{\bbs}_{\ccalL}^{(q)} \|_1 = \| \bbR^\dag \bbR ( \bbs^*_{\ccalL} - \hat{\bbs}_{\ccalL}^{(q)}) \|_1 \leq \| \bbR^\dag \|_{1} \| \bbrho \|_1$.

%%%%%% PROOF OF CLAIM 2 %%%%%%%%%%%%%%%%%%%%%%%%%%%%%
\subsection{Proof of Claim~\ref{C:reqs_3}} 

Since $\textstyle\sum_{j=1}^{N} {S^{(1)}_{j1}}^* = 1$
, then $(\bbe_1 \otimes \bbone_N)^\top \bbs^* = 1$. 
%Since ${\bbS^{(k)}}^* \in \ccalS$
%If we let $\textstyle\sum_{j=1}^{N} S^{(k)}_{j1} = 1$ for all $k$, 
%, we have that $(\bbe_1 \otimes \bbone_N)^\top \bbs^* = \mathbf{1}$. 
%$(\bbI_K \otimes \bbe_1 \otimes \bbone_p)^\top \bbs^* = \mathbf{1}$. 
Thus, having that $ \| \bbPhi_q \bbs^*_{\ccalL} -  \bbb_q \|_2 \leq \epsilon_n $ is equivalent to $\|\bbM \bbs^*_{\ccalL}\|_2 \leq \epsilon_n$ for all $q$. 
From \eqref{E:l1_norm} this is equivalent to requiring that $\sum_{k=1}^K \| {\bbS^{(k)}}^* \hat{\bbC}^{(k)} - \hat{\bbC}^{(k)} {\bbS^{(k)}}^* \|_{\mathrm{F}}^2 \leq \epsilon_n^2$. Hence, we will show that this inequality holds with probability at least $1 - e^{- C' \log N}$ for some constant $C'>0$ when conditions \emph{2)}-\emph{5)} in Theorem~\ref{T:robust_recovery} hold.

Begin by noting that condition \emph{4)} in particular implies that $\log N = o(n)$. This will be used throughout the proof.
Leveraging the fact that ${\bbS^{(k)}}^* {\bbC}^{(k)} = {\bbC}^{(k)} {\bbS^{(k)}}^*$, and making use of the well-known inequality $(a + b)^2 \leq 2 a^2 + 2 b^2$, we denote by $\bbT^{(k)} = {\bbS^{(k)}}^* \hat{\bbC}^{(k)} - \hat{\bbC}^{(k)} {\bbS^{(k)}}^*$, $\bbT^{(k)}_1 = {\bbS^{(k)}}^* \hat{\bbC}^{(k)} - {\bbS^{(k)}}^* \bbC^{(k)}$, and $\bbT^{(k)}_2 =  \bbC^{(k)}  {\bbS^{(k)}}^* - \hat{\bbC}^{(k)} {\bbS^{(k)}}^*$ so that 
\begin{equation*}
\begin{aligned}
\left| [\bbT^{(k)}]_{ij} \right|^2 \leq 2 \left| [\bbT^{(k)}_1]_{ij} \right|^2 + 2 \left| [\bbT^{(k)}_2]_{ij} \right|^2
%&\left| \left[ {\bbS^{(k)}}^* \hat{\bbC}^{(k)} - \hat{\bbC}^{(k)} {\bbS^{(k)}}^* \right]_{ij} \right|^2 \leq 2 \left| \left[ {\bbS^{(k)}}^* \hat{\bbC}^{(k)} - {\bbS^{(k)}}^* \bbC^{(k)} \right]_{ij} \right|^2 \\
%&\qquad + 2 \left| \left[ \hat{\bbC}^{(k)} {\bbS^{(k)}}^* - \bbC^{(k)}  {\bbS^{(k)}}^* \right]_{ij} \right|^2,
\end{aligned}
\end{equation*}
for all $i,j$. In terms of the Frobenius norm of interest, this implies that
\begin{equation}\label{E:proof_joint_1}
\begin{aligned}
& \sum_{k=1}^K \| \bbT^{(k)} \|_{\mathrm{F}}^2 \leq \, N^2 \, \max_{i,j} \sum_{k=1}^K \left| \left[ \bbT^{(k)} \right]_{ij} \right|^2 \\
&  \leq \, 2 N^2 \max_{i,j} \sum_{k=1}^K \left| \left[ \bbT^{(k)}_1 \right]_{ij} \right|^2 + 2 N^2 \max_{i,j} \sum_{k=1}^K \left| \left[ \bbT^{(k)}_2 \right]_{ij} \right|^2.
%& \sum_{k=1}^K \| {\bbS^{(k)}}^* \hat{\bbC}^{(k)} - \hat{\bbC}^{(k)} {\bbS^{(k)}}^* \|_{\mathrm{F}}^2 \\
%& \quad \leq N^2 \, \max_{i,j} \sum_{k=1}^K \left| \left[ {\bbS^{(k)}}^* \hat{\bbC}^{(k)} - \hat{\bbC}^{(k)} {\bbS^{(k)}}^* \right]_{ij} \right|^2 \\
%& \quad \leq 2 N^2 \max_{i,j} \sum_{k=1}^K \left| \left[ {\bbS^{(k)}}^* \hat{\bbC}^{(k)} - {\bbS^{(k)}}^* \bbC^{(k)} \right]_{ij} \right|^2 \\
%& \qquad + 
%2 N^2 \max_{i,j} \sum_{k=1}^K \left| \left[ \hat{\bbC}^{(k)} {\bbS^{(k)}}^* - \bbC^{(k)}  {\bbS^{(k)}}^* \right]_{ij} \right|^2.
\end{aligned}
\end{equation}

We now focus bounding $\max_{i,j} \sum_{k=1}^K \left| \left[ \bbT^{(k)}_1 \right]_{ij} \right|^2$. 
%We now focus bounding $\max_{i,j} \sum_{k=1}^K \left| \left[ {\bbS^{(k)}}^* \hat{\bbC}^{(k)} - {\bbS^{(k)}}^* \bbC^{(k)} \right]_{ij} \right|^2$. 
This is sufficient, since an analogous procedure can be followed to bound the second summand in~\eqref{E:proof_joint_1}. 
In order to bound the first summand in~\eqref{E:proof_joint_1}, we are going to show that the random event
\begin{equation*}
\begin{aligned}
&A := \left\lbrace \sum_{k=1}^K \left| \left[ \bbT^{(k)}_1 \right]_{ij} \right|^2 \leq c_\epsilon^2 \omega^2 K \frac{\log N}{n}, \quad \forall\; i,j\right\rbrace \\
%&A := \left\lbrace \sum_{k=1}^K \left| \left[{\bbS^{(k)}}^* \hat{\bbC}^{(k)} - {\bbS^{(k)}}^* \bbC^{(k)}\right]_{ij} \right|^2 \leq C, \quad \forall\; i,j\right\rbrace \\
%&\qquad \text{where} \,\,\, C = c_\epsilon^2 \omega^2 K \frac{\log N}{n}
\end{aligned}
\end{equation*}
holds with high probability for some constant $c_\epsilon > 0$. Notice that we can regard event $A$ as the intersection of events specific to the entries $(i,j)$, and consider the events
\begin{equation*}
\begin{aligned}
&A_{ij} := \left\lbrace \sum_{k=1}^K \left| \left[ \bbT^{(k)}_1 \right]_{ij} \right|^2 \leq c_\epsilon^2 \omega^2 K \frac{\log N}{n}\right\rbrace. \\
\end{aligned}
\end{equation*}

Recall that $\bbX^{(k)} \in \reals^{N \times n_k}$ contains the signals $\bbx^{(k)}_i$ as columns. We denote as $(\bbz_j^{(k)})^\top \in \reals^{n_k}$ the $j$-th row of $\bbX^{(k)}$, i.e., the vector collecting the value in the $j$-th position of each of the graph signals associated with the $k$-th GSO. Moreover, for simplicity we will denote by $(\bbs_i^{(k)})^\top \in \reals^N$ the $i$-th row of ${\bbS^{(k)}}^*$. From~\eqref{E:sample_covariance_k}, we then have that
%Notice that $\hat{\bbC}^{(k)} = 1/n_k (\bbX^{(k)})^\top \bbX^{(k)}$ where the matrix $\bbX^{(k)} \in \reals^{n_k \times N}$ contains the signals $(\bbx^{(k)}_i)^\top$ as rows. We denote as $\bbz_j^{(k)} \in \reals^{n_k}$ the $j$-th column of $\bbX^{(k)}$, i.e., the vector collecting the value in the $j$-th position of each of the graph signals associated with the $k$-th GSO. Moreover, for simplicity we will denote by $(\bbs_i^{(k)})^\top \in \reals^N$ the $i$-th row of ${\bbS^{(k)}}^*$. We then have that
%
\begin{equation}\label{E:proof_joint_2}
\left| \left[ \bbT^{(k)}_1 \right]_{ij} \right| = \frac{1}{n_k} \left| (\bbs_i^{(k)})^\top \bbX^{(k)} (\bbz_j^{(k)}) -  \mathbb{E}[(\bbs_i^{(k)})^\top \bbX^{(k)} (\bbz_j^{(k)})]\right|,
%\left| \left[ \bbT^{(k)}_1 \right]_{ij} \right| = \frac{1}{n_k} \left| (\bbs_i^{(k)})^\top (\bbX^{(k)})^\top \bbz_j^{(k)} -  \mathbb{E}[(\bbs_i^{(k)})^\top (\bbX^{(k)})^\top \bbz_j^{(k)}]\right|,
%\left| \left[{\bbS^{(k)}}^* \hat{\bbC}^{(k)} - {\bbS^{(k)}}^* \bbC^{(k)} \right]_{ij} \right| = \frac{1}{n_k} \left| (\bbs_i^{(k)})^\top (\bbX^{(k)})^\top \bbz_j^{(k)} -  \mathbb{E}[(\bbs_i^{(k)})^\top (\bbX^{(k)})^\top \bbz_j^{(k)}]\right|,
\end{equation}
where, under a slight abuse of notation, we are now considering $\bbX^{(k)}$ and $\bbz_j^{(k)}$ as random variables instead of specific realizations. Given that the columns of $\bbX^{(k)}$ are i.i.d., we have that $(\bby_i^{(k)})^\top := (\bbs_i^{(k)})^\top \bbX^{(k)} \sim \ccalN(\mathbf{0}, (\bbs_i^{(k)})^\top \bbC^{(k)} \bbs_i^{(k)} \bbI)$ and, by definition, $(\bbz_j ^{(k)})^\top\sim \ccalN( \mathbf{0}, [\bbC^{(k)}]_{jj} \bbI)$. 
It then follows that each term in the sum $(\bby_i^{(k)})^\top (\bbz_j^{(k)}) = \textstyle\sum_{t = 1}^{n_k} (\bby_i^{(k)})_t (\bbz_j^{(k)})_t$, is i.i.d. Leveraging this decomposition, we may write
\begin{equation*}
\left| \left[ \bbT^{(k)}_1 \right]_{ij} \right| = \left| \frac{1}{n_k} \sum_{t=1}^{n_k} y_{i,t}^{(k)} z_{j,t}^{(k)} -  \mathbb{E}[ y_i^{(k)} z_j^{(k)}]\right|,
%\left| \left[{\bbS^{(k)}}^* \hat{\bbC}^{(k)} - {\bbS^{(k)}}^* \bbC^{(k)} \right]_{ij} \right| = \left| \frac{1}{n_k} \sum_{t=1}^{n_k} y_{i,t}^{(k)} z_{j,t}^{(k)} -  \mathbb{E}[ y_i^{(k)} z_j^{(k)}]\right|,
\end{equation*}
where we denote by $y_i^{(k)}$ a scalar random variable representing the elements of $\bby_i^{(k)}$ (since they are all i.i.d.) and as $y_{i,t}^{(k)}$ a specific realization of this random variable. 
The same applies for $z_j^{(k)}$ with respect to $\bbz_j^{(k)}$. 
We denote random variables $w_{i+j}^{(k)} = y_i^{(k)} + z_j^{(k)}$ and $w_{i-j}^{(k)} = y_i^{(k)} - z_j^{(k)}$. By subsequently applying the identity $4ab = (a+b)^2 - (a-b)^2$ and the inequality $(a+b)^2 \leq 2a^2 + 2b^2$ we obtain that
% %
% \begin{equation*}
% \begin{split}
% \left| \left[ \bbT^{(k)}_1 \right]_{ij} \right|  = &\frac{1}{4} \left|
% \frac{1}{n_k} \sum_{t=1}^{n_k} (w_{i+j,t}^{(k)})^2 -  \mathbb{E}[ (w_{i+j}^{(k)})^2 ] \right. \\
% &\left. - \frac{1}{n_k} \sum_{t=1}^{n_k} (w_{i-j,t}^{(k)})^2 +  \mathbb{E}[ (w_{i-j}^{(k)})^2 ] \vphantom{\frac{1}{n_k} \sum_{t=1}^{n_k}} \right|.
% %\left| \left[{\bbS^{(k)}}^* \hat{\bbC}^{(k)} - {\bbS^{(k)}}^* \bbC^{(k)} \right]_{ij} \right|  = \frac{1}{4} \left|
% %\frac{1}{n_k} \sum_{t=1}^{n_k} (y_{i,t}^{(k)} + z_{j,t}^{(k)})^2 -  \mathbb{E}[ (y_i^{(k)} + z_j^{(k)})^2 ]
% %+ \frac{1}{n_k} \sum_{t=1}^{n_k} (y_{i,t}^{(k)} - z_{j,t}^{(k)})^2 -  \mathbb{E}[ (y_i^{(k)} - z_j^{(k)})^2 ] \right|.
% \end{split}
% \end{equation*}
% %
% %
% Moreover, leveraging the inequality $(a+b)^2 \leq 2a^2 + 2b^2$ we obtain that
%
\begin{equation} \label{E:proof_joint_10}
\begin{split}
\left| \left[ \bbT^{(k)}_1 \right]_{ij} \right|^2 \leq &\frac{1}{8} \left|
\frac{1}{n_k} \sum_{t=1}^{n_k} (w_{i+j,t}^{(k)})^2 -  \mathbb{E}[ (w_{i+j}^{(k)})^2 ] \right|^2 \\
& + \frac{1}{8}  \left| \frac{1}{n_k} \sum_{t=1}^{n_k} (w_{i-j,t}^{(k)})^2 -  \mathbb{E}[ (w_{i-j}^{(k)})^2 ] \right|^2.
\end{split}
\end{equation}
Observe that, since both $y_i^{(k)}$ and $z_j^{(k)}$ are Gaussian random variables, we have that $w_{i+j}^{(k)}$ and $w_{i-j}^{(k)}$ are also Gaussian with variance at most $4 \omega_k$. 
We define $\rho_k := n_k / n$ and, for fixed $i,j$, we define $u_t^{(k)} := (\rho_k^{1/2}/n_k) \big( (w_{i+j,t}^{(k)})^2 -  \mathbb{E}[ (w_{i+j}^{(k)})^2] \big)$ if $t \leq n_k$ and $u_t^{(k)} = 0$ if $t > n_k$. Also, let $\bbu_t := (u_t^{(1)}, \cdots, u_t^{(K)})^\top$. By definition, we can then write
\begin{equation*}
\sum_{k=1}^K \rho_k \left( \left| \frac{1}{n_k} \sum_{t=1}^{n_k} (w_{i+j,t}^{(k)})^2 - \mathbb{E}[ (w_{i+j}^{(k)})^2] \right| \right)^2 = \left\Vert \sum_{t=1}^n \bbu_t \right\Vert_2^2.
\end{equation*}
Consider now a new event $A'_{ij}$ based on the newly introduced variable $\bbu_t$, namely
$$
A_{ij}' := \left\lbrace \left\Vert \sum_{t=1}^n \bbu_t \right\Vert_2^2 \leq c'^2_\epsilon \omega^2 {\frac{\log N}{n}} \right\rbrace,
$$
for some constant $c'_\epsilon$. We now briefly show that there exists a constant $c'_\epsilon$ such that the probability of $A_{ij}'$ occurring is not larger than the probability of occurrence of $A_{ij}$. Indeed, from condition \emph{3)} we know that there exists some constant $c_w$ such that $\rho_k \geq c_w / K$. Hence, when $A_{ij}'$ occurs, it is also satisfied that
\begin{equation} \label{E:proof_joint_3}
\sum_{k=1}^K \left| \frac{1}{n_k} \sum_{t=1}^{n_k} (w_{i+j,t}^{(k)})^2 - \mathbb{E}[ (w_{i+j}^{(k)})^2] \right|^2 \leq (c'^2_\epsilon / c_w) \omega^2 K \frac{\log N}{n}.
\end{equation}
A similar analysis can be used to bound the above expression but for $w_{i-j}^{(k)}$ instead of $w_{i+j}^{(k)}$. 
Hence, we substitute these bounds in the expression obtained by summing~\eqref{E:proof_joint_10} over all $k=1, \ldots, K$, to see that $A_{ij}$ also occurs, where the constant $c_\epsilon$ in $A_{ij}$ depends on $c'_\epsilon$ and $c_w$.
Consequently, if we show that $\mathbb{P}(A'_{ij}) \geq 1 - c' e^{-c \log N}$ for some constants $c>2$ and $c'$, it would then follow that $\mathbb{P} (A_{ij}) \geq 1 - c' e^{-c \log N}$. Moreover, a union bound over all $(i,j)$ then guarantees the existence of a constant $C'>0$ such that $\mathbb{P} (A) \geq 1 - e^{-C' \log N}$. It follows from the discussion after \eqref{E:proof_joint_1} that this would complete the proof.

Consequently, we are left to show that under conditions \emph{2)}-\emph{5)} we have that $\mathbb{P} (A'_{ij}) \geq 1 - c' e^{-c \log N}$ for some constants $c>2$ and $c'$. 
The remainder of the proof of Claim~\ref{C:reqs_3} is devoted to proving this statement. We are going to prove this by showing that $\mathbb{P}(\neg A'_{ij}) \leq c' e^{-c \log N}$.

Notice that we cannot directly use Lemma~\ref{L:prob_bound_norm} to bound $\| \sum_{t=1}^n \bbu_t \|^2_2$, since we would need $\|\bbu_t\|_2$ to be bounded by some constant $M$. We therefore split $A_{ij}'$ into two subevents and estimate the bound for the probability of each of the two subevents. The basic intuition is that, if we are on the random event 
$$E_a := \left\lbrace \vert u_t^{(k)} \vert \leq (n \log N)^{-1/2} K^{1/2-a} \;\omega,\;\,\,\, \forall\; t, k\right\rbrace,$$
then the $\ell_2$ norm of $\bbu_t$ would always be smaller than 
\begin{equation}\label{E:proof_joint_9}
\| \bbu_t \|_2 \leq M_a := (n \log N)^{-1/2} K^{1-a} \;\omega,
\end{equation}
where $a$ is a free parameter that will be fixed later in the proof.

In particular, if we split the complement of $A_{ij}'$ as 
\begin{equation}\label{E:bound_neg_A_ij}
\mathbb{P}(\neg A_{ij}') \leq \mathbb{P}(\neg A_{ij}' \vert E_a) \mathbb{P}(E_a) + \mathbb{P}(\neg E_a),
\end{equation}
we can use Lemma~\ref{L:prob_bound_norm} to bound $\mathbb{P}(\neg A_{ij}' \vert E_a)$ and then use Lemma~\ref{L:tail_bounds} to bound $\mathbb{P}(\neg E_a)$.
Let us introduce a new variable
\begin{equation*}
\begin{split}
\hat{u}_t^{(k)} := &\, u_t^{(k)} I\left\lbrace \vert u_t^{(k)} \vert \leq (n \log N)^{-1/2} K^{1/2-a} \omega \right\rbrace \\ 
& - \mathbb{E}\left[ u_t^{(k)} I\left\lbrace \vert u_t^{(k)} \vert \leq (n \log N)^{-1/2} K^{1/2-a} \omega \right\rbrace \right]
\end{split}
\end{equation*}
and $\hat{\bbu}_t := (\hat{u}_t^{(1)}, \cdots, \hat{u}_t^{(K)})^\top$. 
Notice that if we are on the random event $E_a$, then $\bbu_t$ and $\hat{\bbu}_t$ follow the same distribution except for a shift $v_t^{(k)} := \mathbb{E}\left[ u_t^{(k)} I\left\lbrace \vert u_t^{(k)} \vert \leq (n \log N)^{-1/2} K^{1/2-a} \omega \right\rbrace \right]$. 
Putting it differently, the distribution of $u_t^{(k)} - \hat{u}_t^{(k)}$ is a constant $v_t^{(k)}$ when we are on the random event $E_a$. 

We can use Lemma~\ref{L:tail_bounds} to estimate the scale of $v_t^{(k)}$ with respect to $N$ and $n$. To do this, first notice that $u_t^{(k)}$ is a chi-squared random variable with one degree of freedom. Thus, for some constant $\eta$ we can apply Lemma~\ref{L:tail_bounds} for $\sigma^2 = \frac{\sqrt{K} \omega}{8 \eta n}$, $t =  \frac{8 \eta n l}{\sqrt{K} \omega}$ and $m = 1$, to obtain the tail bound
\begin{equation}\label{E:tail_bound_specific_chi}
\mathbb{P}(u_t^{(k)} \geq l) \leq \exp\left(- \eta \frac{n l}{\sqrt{K} \omega} \right) \quad \textrm{with}\quad l \gg \frac{\sqrt{K} \omega}{n}.
\end{equation}
Moreover, since $u_t^{(k)}$ has mean zero for all $t$ and $k$, we have that
\begin{equation}\label{E:proof_joint_5}
\begin{aligned}
&\vert v_t^{(k)} \vert  &=& \left\vert \mathbb{E}\left[ u_t^{(k)} I\left\lbrace \vert u_t^{(k)} \vert \leq (n \log N)^{-1/2} K^{1/2-a} \omega \right\rbrace \right] \right\vert \\
& &=& \left\vert \mathbb{E}\left[ u_t^{(k)} I\left\lbrace \vert u_t^{(k)} \vert \geq (n \log N)^{-1/2} K^{1/2-a} \omega \right\rbrace \right] \right\vert.
\end{aligned}
\end{equation}
It follows from the definition of $u_t^{(k)}$ that $u_t^{(k)} \geq - \rho_k^{1/2} (4 \omega/n_k)$. 
From the fact that $\log N = o(n)$, we have that 
\begin{equation} \label{E:proof_joint_8}
\rho_k^{1/2} \frac{4 \omega}{n_k} << (n \log N)^{-1/2} K^{1/2-a} \omega.
\end{equation}
Therefore, combining both previous facts, when $u_t^{(k)}$ satisfies that $\vert u_t^{(k)} \vert \geq (n \log N)^{-1/2} K^{1/2-a} \omega$, it must be that $u_t^{(k)}$ is positive. 
Therefore, the right hand side of~\eqref{E:proof_joint_5} can be further rewritten as $\mathbb{E}\left[ u_t^{(k)} I\left\lbrace u_t^{(k)} \geq (n \log N)^{-1/2} K^{1/2-a} \omega \right\rbrace \right]$, where we have deleted the absolute value of $u_t^{(k)}$. 
We let $\theta := (n \log N)^{-1/2} K^{1/2-a} \omega$ and $\gamma := \eta \frac{n}{\sqrt{K} \omega}$.
Consequently, we may bound $\vert v_t^{(k)} \vert$ as follows
\begin{equation*}
\begin{split}
\vert v_t^{(k)} \vert & = \mathbb{E}\left[ u_t^{(k)} I\left\lbrace u_t^{(k)} \geq (n \log N)^{-1/2} K^{1/2-a} \omega \right\rbrace \right] \\
& \leq \int_{\theta}^{\infty} \exp(-\gamma \ell) \, d \ell 
= \frac{1}{\gamma} \exp(-\gamma \theta) \\
& = \frac{\sqrt{K} \omega}{\eta n} \exp \left( - \eta (n / \log N)^{1/2} K^{-a} \right),
\end{split}
\end{equation*}
where we have used~\eqref{E:tail_bound_specific_chi} in the computation of the expected value. 
Clearly, $\vert v_t^{(k)} \vert$ decays exponentially with respect to $n / \log N$. 
Therefore, we have that $n \vert v_t^{(k)} \vert = o(\sqrt{(\log N)/n})$ for all $k = 1, \ldots, K$. In addition, when we are on the random event $E_a$, we have that $\sum_{t=1}^n u_t^{(k)} = \sum_{t=1}^n \hat{u}_t^{(k)} + n_k v_t^{(k)}$ and therefore $(\sum_{t=1}^n u_t^{(k)})^2/2 \leq (\sum_{t=1}^n \hat{u}_t^{(k)})^2 + (n_k v_t^{(k)})^2$. By summing the previous expression over all $k = 1, \ldots, K$, we further have that
\begin{equation} \label{E:proof_joint_7}
\frac{1}{2}\left\Vert \sum_{t=1}^n \bbu_t \right\Vert_2^2 \leq  \sum_{k=1}^K (n_k \, v_t^{(k)})^2 + \left\Vert \sum_{t=1}^n \hat{\bbu}_t \right\Vert_2^2.
\end{equation} 
By combining~\eqref{E:proof_joint_7} with the fact that $n \vert v_t^{(k)} \vert = o(\sqrt{(\log N)/n})$ for all $k$, we further have that there exists some $0 < \delta < 1$ such that if we are on the event $E_a$, then $\Vert \sum_{t=1}^n \bbu_t \Vert_2 \geq c'_\epsilon \omega \sqrt{\frac{\log N}{n}}$ indicates that $\Vert \sum_{t=1}^n \hat{\bbu}_t \Vert_2 \geq (1 - \delta) c'_\epsilon \omega \sqrt{\frac{\log N}{n}}$.
Equivalently, there exists some constant $0 < \delta < 1$ such that, given the event,
\begin{equation*}
B = \left\{ \left\| \sum_{t=1}^n \hat{\bbu}_t \right\|_2 \geq (1 - \delta) c'_\epsilon \omega \sqrt{\frac{\log N}{n}} \right\}, 
\end{equation*}
the following inequality holds 
\begin{equation}\label{E:proof_joint_4}
\mathbb{P}(\neg A_{ij}' \vert E_a) \mathbb{P}(E_a) \leq \mathbb{P} ( B\;  \vert \; E_a ) \mathbb{P}(E_a) \leq \mathbb{P}(B),
%\mathbb{P}(\neg A_{ij}' \vert E_a) \leq \mathbb{P} \left( \left\| \sum_{t=1}^n \hat{\bbu}_t \right\|_2 \geq (1 - \delta) c'_\epsilon \omega \sqrt{\frac{\log N}{n}}\;  \bigg\vert \; E_a \right),
%\mathbb{P}(\neg A_{ij}' \vert E_a) \mathbb{P}(E_a) \leq \mathbb{P} \left( \left\| \sum_{t=1}^n \hat{\bbu}_t \right\|_2 \geq (1 - \delta) c'_\epsilon \omega \sqrt{\frac{\log N}{n}}\;  \bigg\vert \; E_a \right) \mathbb{P}(E_a) \leq \mathbb{P} \left( \left\| \sum_{t=1}^n \hat{\bbu}_t \right\|_2 \geq (1 - \delta) c'_\epsilon \omega \sqrt{\frac{\log N}{n}} \right),
\end{equation}
where the second inequality follows readily from Bayes' theorem. 
Given that $\hat{\bbu}_t$ is obtained from $\bbu_t$ by cutting the tails, we have that $\lambda_{\max} \left\lbrace \textrm{Cov}(\sum_{t=1}^n \hat{\bbu}_t) \right\rbrace \leq \lambda_{\max} \left\lbrace \textrm{Cov}(\sum_{t=1}^n \bbu_t) \right\rbrace$. In addition, as $u_t^{(k)}$ are i.i.d. for all $t$, we have that 
\begin{equation*}
\begin{aligned}
&\lambda_{\max} \left\lbrace \textrm{Cov}\left(\sum_{t=1}^n \hat{\bbu}_t \right) \right\rbrace = \max_k \textrm{Var}\left(\sum_{t=1}^n u_t^{(k)} \right) \\
& \qquad \qquad = (n_k \rho_k / n^2_k) \textrm{Var}\left((w_{i+j}^{(k)})^2\right) \leq 16 \omega^2 / n.
\end{aligned}
\end{equation*}
We may thus apply Lemma~\ref{L:prob_bound_norm} to further bound \eqref{E:proof_joint_4} by setting $r = (1 - \delta) c'_\epsilon \omega \sqrt{\frac{\log N}{n}}$, $s = \frac{1}{2} (1 - \delta) c'_\epsilon \omega \sqrt{\frac{\log N}{n}}$, $\lambda_{\max} = 16 \omega^2 / n$ and $M = M_a$ as defined in~\eqref{E:proof_joint_9} to get that
\begin{equation}\label{E:proof_multi_100}
\begin{aligned}
	&\mathbb{P} \left(B \right) \leq&& \mathbb{P}( \Vert \bbz \Vert_2 \geq C_2 \sqrt{\log N}) \\ & &&+ C_3 \exp \left\lbrace \frac{2}{5} \log K - C_4 K^{a-7/2} \log N\right\rbrace
\end{aligned}
\end{equation}
% \begin{equation}\label{E:proof_multi_100}
% \begin{aligned}
% 	&\mathbb{P} \left(B \right) \leq&& \mathbb{P}( \Vert \bbz \Vert_2 \geq C_2 \sqrt{\log N}) \\ & &&+ C_3 \exp \left\lbrace \frac{2}{5} \log K - C_4 K^{a-\frac{7}{2}} \log N\right\rbrace
% \end{aligned}
% \end{equation}
%
where $C_2$ and $C_4$ are constants that increase with increasing $c_\epsilon$ and $\bbz$ is a $K$-dimensional standard normal random vector. Note that the constant $\delta$ is also absorbed into $C_2$ and $C_4$. From condition \emph{2)} and the tail bound in Lemma~\ref{L:tail_bounds} we get that
\begin{equation*}
\mathbb{P}( \Vert \bbz \Vert_2 \geq C_2 \sqrt{\log N}) \leq C_1 \exp(-C_2' (\log N - K))
\end{equation*}
where $C_2'$ is a constant that increases with increasing of $c_\epsilon$. 
Replacing the above expression into \eqref{E:proof_multi_100} we obtain 
\begin{equation*}
\begin{aligned}
\mathbb{P} \!\left( B \right) \!\leq \!\exp \lbrace - C_2' \log N  \rbrace + C_3 \exp \left\lbrace \!\frac{2 \log K}{5}  - \frac{C_4 \log N}{K^{7/2-a}} \!\right\rbrace
\end{aligned}
\end{equation*}
% \begin{equation*}
% \begin{aligned}
% \mathbb{P} \!\left( B \right) \!\leq \!\exp \lbrace - C_2' \log N  \rbrace + C_3 \exp \left\lbrace \!\frac{2 \log K}{5}  - \frac{C_4 \log N}{K^{\frac{7}{2}-a}} \!\right\rbrace
% \end{aligned}
% \end{equation*}
%
where $C_2'$ and $C_4$ are the constants that would increase with the increment of $c_\epsilon$. In this case, by choosing $a = 7/2$, with constant $c_\epsilon$ big enough, there exists some constant $c_1 > 2$ such that 
\begin{equation}\label{E:proof_multi_110}
\mathbb{P} \left( B \right) \leq \exp(-c_1 \log N).
\end{equation}

Replacing \eqref{E:proof_multi_110} into \eqref{E:proof_joint_4} gives us the sought exponential bound for the first summand in \eqref{E:bound_neg_A_ij}. We are now left with the task of finding a bound for $\mathbb{P}(\neg E_a)$.

Given event $B_{(k)}' = \left\{\vert  u_t^{(k)}\vert \geq \left( \frac{K^{1 - 2a}}{n \log N} \right)^{1 / 2} \omega \right\}$, from the definition of the event $E_a$ it follows that 
\begin{equation}\label{E:proof_joint_6}
%\mathbb{P}(\neg E_a) \leq K \left( \max_{1 \leq k \leq K} n_k \right) \max_{1 \leq k \leq K} \; \mathbb{P} \left( \vert  u_t^{(k)}\vert \geq \left( \frac{K^{1 - 2a}}{n \log N} \right)^{1 / 2} \omega \right).
\mathbb{P}(\neg E_a) \leq K \left( \max_{1 \leq k \leq K} n_k \right) \max_{1 \leq k \leq K} \; \mathbb{P} \left( B_{(k)}' \right).
\end{equation}
From~\eqref{E:proof_joint_8} we obtain that the probabilities $\mathbb{P}\left( u_t^{(k)} \geq \left( \frac{K^{1 - 2a}}{n \log N} \right)^{1 / 2} \omega \right)$ and $\mathbb{P}\left( \vert u_t^{(k)} \vert \geq \left( \frac{K^{1 - 2a}}{n \log N} \right)^{1 / 2} \omega \right)$ are the same. 
Plus, from the tail bound in~\eqref{E:tail_bound_specific_chi} we get
\begin{equation*}
\begin{aligned}
\mathbb{P} \left( B_{(k)}' \right) \leq e^{ - \eta \sqrt{ \frac{n}{K^{2a} \log N}} } = e^{ - \eta \sqrt{\frac{n}{K^{7} \log N}} },
\end{aligned}
\end{equation*}
where the last equality follows from recalling that we have fixed $a = 7/2$ in order to write \eqref{E:proof_multi_110}. Condition \emph{3)} guarantees the existence of some constant $c_w$ such that $n_k \leq n c_w / K$ for all $k$, thus
\begin{equation}\label{E:proof_multi_120}
\begin{aligned}
\mathbb{P}(\neg E_a) \leq c_w n e^{ - \eta \sqrt{ \frac{n}{K^{7} \log N}}}
\leq c_w e^{ \log n - \eta \sqrt{\frac{n}{K^{7} \log N}}}.
\end{aligned}
\end{equation}
From condition \emph{4)} it follows that $\log n = o( \sqrt{ n/ (K^7 \log N)})$ and $\log N = o( \sqrt{ n/ (K^7 \log N)})$, which immediately implies that $\log N = o\left(\sqrt{ \frac{n}{K^7 \log N} } - \log n \right)$.
Combining this expression with \eqref{E:proof_multi_120} reveals that 
\begin{equation}\label{E:proof_multi_130}
\mathbb{P}(\neg E_a) \leq c_w  e^{-c_2 \log N},
\end{equation}
for some constant $c_2>2$, thus obtaining an exponential bound for the second summand in \eqref{E:bound_neg_A_ij}. To conclude, from the combination of \eqref{E:proof_multi_110} and \eqref{E:proof_multi_130} we get that $\mathbb{P}(\neg A_{ij}') \leq c' \exp(-c \log N)$ for some $c > 2$, as wanted.
%%%%%%%%%%%%%%%%%%%%%%

%%%%%%%%%%%%%%%%%%%%%%%%%%%%%%%%%%%%%%%%%%%%%%%%%%%%%%%%%%%%%%
% BIBLIOGRAPHY
\bibliographystyle{IEEEtran}
%% argument is your BibTeX string definitions and bibliography database(s)
%
\bibliography{citations}

\end{document}